\pdfoutput=1
\documentclass[11pt,letterpaper]{article}

\usepackage[english]{babel}
\usepackage[utf8]{inputenc}
\usepackage[T1]{fontenc}
\usepackage{lmodern}
\usepackage[margin=1in]{geometry}
\usepackage{amsmath,amsthm,amssymb,mathtools}
\usepackage{microtype}
\usepackage{graphicx}
\usepackage{subcaption}
\usepackage{booktabs}
\usepackage{multirow}
\usepackage{enumitem}
\usepackage{xcolor}
\usepackage[numbers,sort&compress]{natbib}
\usepackage{url}
\usepackage[colorlinks=true,linkcolor=blue!50!black,citecolor=blue!50!black,urlcolor=blue!50!black]{hyperref}

\hypersetup{
  pdftitle={From Theory to Decision Rule: Calibrating the Noisy-Label Crossover for Vision-Language Model Weak Supervision Across Three Medical-Imaging Benchmarks},
  pdfauthor={Bruce Changlong Xu, Jose K. James, Alexander J. Ryu},
  pdfsubject={Machine Learning (cs.LG); Computer Vision and Pattern Recognition (cs.CV)}
}

\theoremstyle{plain}
\newtheorem{theorem}{Theorem}[section]
\newtheorem{proposition}[theorem]{Proposition}

\theoremstyle{definition}
\newtheorem{definition}[theorem]{Definition}

\newcommand{\ngold}{n_g}
\newcommand{\nweak}{n_w}
\newcommand{\lamb}{\lambda}
\newcommand{\muS}{$\mu$-scaling}

\title{\vspace{-1.4em}\textbf{From Theory to Decision Rule: Calibrating the Noisy-Label Crossover for Vision-Language Model Weak Supervision Across Three Medical-Imaging Benchmarks}\thanks{Accepted at the ICML 2026 Workshop on Combining Theory and Benchmarks (CTB). This extended preprint includes the full theoretical derivation of the bias--variance noisy-label crossover, per-pathology heterogeneity analysis across all 15 NIH ChestX-ray14 classes, uncertainty-aware decision-rule audits, and camera-ready prompt, loss, and cross-VLM robustness sweeps.}}

\author{
  \textbf{Bruce Changlong Xu}\textsuperscript{1}\thanks{Corresponding authors: \texttt{brucechanglongxu@cs.stanford.edu}, \texttt{Ryu.Alexander@mayo.edu}.} \qquad
  \textbf{Jose K. James}\textsuperscript{2} \qquad
  \textbf{Alexander J. Ryu}\textsuperscript{2}\footnotemark[2] \\[4pt]
  \textsuperscript{1}Department of Computer Science, Stanford University, Stanford, CA, USA \\
  \textsuperscript{2}Mayo Clinic, Rochester, MN, USA
}

\date{\vspace{-1.0em}}

\begin{document}

\maketitle
\thispagestyle{plain}

\begin{abstract}
\noindent
Classical noisy-label learning theory establishes that downstream risk under corrupted supervision is bounded below by an error floor governed by the labeler's noise structure, implying a sharp \emph{crossover} along the gold-annotation budget axis: once a classifier trained solely on expert gold labels matches the weak labeler's effective accuracy ceiling, adding weak pseudo-labels stops reducing error and begins degrading test performance. Translating this theoretical phase transition into an operational rule for modern vision--language foundation models (VLMs) requires empirical calibration across tasks, student architectures, and noise structures. We provide a theoretical and empirical calibration for VLM weak supervision across three clinical imaging benchmarks---PatchCamelyon (PCAM) histopathology, ISIC~2019 dermoscopy, and NIH ChestX-ray14 radiography---spanning two biomedical VLMs (BiomedCLIP and PubMedCLIP) and six student architectures covering an $11\times$ parameter range ($8\text{M}$--$89\text{M}$). We observe a sharp crossover at $n_g \approx 100$ on PCAM, $n_g \in [20, 50]$ on ISIC~2019, and $n_g \in [250, 500]$ on NIH ChestX-ray14; above these thresholds, weak-label augmentation degrades macro-AUC by up to $-0.14$ on ISIC with BiomedCLIP ($-0.23$ with PubMedCLIP), $-0.10$ on PCAM, and $-0.08$ on NIH ChestX-ray14. Across the six-architecture PCAM sweep and a within-family DenseNet ladder (DenseNet-121/169/201), the weak-dominated ceiling clusters within a $0.024$ AUC band ($0.841$--$0.865$), confirming that the VLM labeler rather than student capacity is the binding constraint. Comparing BiomedCLIP pseudo-labels to rate-matched symmetric random noise reveals a sign reversal: structured VLM errors underperform random corruption by $0.09$--$0.11$ AUC on PCAM and $0.08$--$0.20$ AUC on ISIC~2019, yet outperform random noise by $+0.03$--$0.04$ AUC on 15-class NIH ChestX-ray14 because coarse normal-versus-abnormal predictions preserve cross-pathology signal, while per-pathology crossovers span $>20\times$ and expose anti-correlated \emph{toxic} classes (\texttt{No Finding} proxy AUC $0.284$). Finally, we distill our bias--variance scaling law into a zero-fitting decision rule operable from $10$--$50$ validation labels---add weak labels only when gold-only validation AUC falls below the VLM's validation ceiling---which predicts the correct sign on $30/30$ statistically significant grid cells ($39/42$ overall) and remains calibrated across prompt, loss, and cross-VLM stress tests.
\end{abstract}

\vspace{6pt}

\section{Introduction}
\label{sec:intro}

A pervasive paradigm for adapting multimodal foundation models to specialized scientific and clinical domains is to deploy the foundation model as a zero-shot \emph{weak labeler}: prompt a large pretrained vision--language model (VLM) to annotate an unlabeled image pool at near-zero marginal cost, combine those pseudo-labels with a small seed set of expert-annotated \emph{gold} examples, and train a compact, task-specific student network on the union \citep{hinton2015distilling,ratner2016data,ratner2017snorkel,xie2020selftraining,sohn2020fixmatch}. In clinical computer vision, this workflow transforms a $400\text{M}$-parameter biomedical VLM such as BiomedCLIP \citep{zhang2024biomedclip}, PubMedCLIP \citep{eslami2023pubmedclip}, CheXzero \citep{tiu2022expert}, PLIP \citep{huang2023visual}, or CONCH \citep{lu2024conch}---all building on contrastive language--image pretraining \citep{radford2021learning}---into an inexpensive supervisory scaffold for a deployable $8\text{M}$-parameter convolutional network \citep{he2016deep,huang2017densely}. Because the foundation model and the downstream student share neither parameters nor gradients, VLM pseudo-labeling functions as a drop-in data-augmentation primitive. In practice, clinical AI pipelines frequently operate under the implicit assumption that expanding the weak-labeled pool is monotonically beneficial, reducing the engineering problem to confidence thresholding or consensus filtering \citep{patrini2017making,northcutt2021confident,zhong2024vlmcpl}.

Classical learning theory under corrupted supervision predicts a markedly different phenomenon \citep{angluin1988learning,natarajan2013learning,frenay2014classification,menon2015learning,ghosh2017robust,song2022learning}. When weak labels are generated by a fixed deterministic teacher $f_{\mathrm{FM}}$ whose predictions exhibit systematic, input-dependent errors, the empirical risk minimizer trained on a mixture of $n_g$ gold labels and $n_w$ weak labels converges as $n_w \to \infty$ not to the Bayes-optimal classifier $f^\star$, but to an interpolant bounded by the teacher's systematic bias floor. Consequently, theory implies a sharp regime transition---a \emph{noisy-label crossover} $n_g^\star$---along the gold-annotation budget axis:
\begin{itemize}[leftmargin=1.8em,itemsep=2pt,topsep=3pt]
  \item \textbf{Below the crossover ($n_g < n_g^\star$):} Gold-only estimation variance exceeds the VLM's systematic bias. Adding VLM pseudo-labels substantially reduces variance and improves downstream generalization.
  \item \textbf{Above the crossover ($n_g > n_g^\star$):} A student trained solely on gold annotations already surpasses the VLM's effective accuracy ceiling. Additional weak labels pull the learned decision boundary back toward the VLM's systematic errors, actively degrading test AUC.
\end{itemize}

Despite the clarity of this qualitative prediction, a critical gap separates noisy-label theory from modern foundation-model practice. Neural scaling laws characterize how test loss scales with dataset size and compute under homogeneous supervision \citep{hestness2017deep,kaplan2020scaling,hoffmann2022training,sorscher2022beyond,cherti2023reproducible,bahri2024explaining}, yet existing scaling formulations do not tell a clinical practitioner \emph{where} the gold-versus-weak crossover occurs for a given biomedical VLM, \emph{whether} scaling up the downstream student architecture can raise the weak-label ceiling, \emph{how} structured VLM confusions differ from classical symmetric label noise, or \emph{how} to decide at deployment time---using only a small validation split---whether to include VLM pseudo-labels at all.

\paragraph{Closing the Theory--Benchmark Loop.}
This work bridges classical noisy-label theory and clinical foundation-model benchmarking through a four-stage investigation:
\begin{enumerate}[leftmargin=1.8em,itemsep=3pt,topsep=3pt]
  \item \textbf{Theoretical bias--variance scaling decomposition and crossover characterization (Section~\ref{sec:theory}).} We formalize mixed gold/weak training via a two-term \emph{micro-scaling} (\muS) law that decomposes downstream test risk into a sample-variance term governed by an effective sample size $n_{\mathrm{eff}} = n_g + \lambda n_w$ and a systematic-bias term $\delta(1 + \gamma n_g)^{-\rho}$ that only gold supervision can eliminate. From this decomposition we derive the closed-form mutual-information predictor $\lambda_{\mathrm{pred}} = I(Y; \widetilde{Y})/H(Y)$, an error-correlation penalty $(1 - \rho_{\mathrm{err}})$ explaining why correlated VLM mistakes violate independent-channel scaling, an explicit characterization of the crossover budget $n_g^\star$, a formal definition of anti-correlated \emph{toxic classes} ($\mathrm{AUC}_k < 0.5$), and a label-space projection proposition explaining when structured VLM noise underperforms versus outperforms matched-rate random noise.
  \item \textbf{Multi-benchmark and multi-architecture empirical calibration (Sections~\ref{sec:setup}--\ref{sec:capacity}).} We instantiate and calibrate the theoretical bound across three public medical-imaging benchmarks spanning distinct clinical modalities and label granularities: \textbf{PatchCamelyon (PCAM)} histopathology \citep{veeling2018rotation,bejnordi2017diagnostic}, \textbf{ISIC~2019} dermoscopy \citep{tschandl2018ham10000,combalia2019bcn20000,codella2018skin}, and \textbf{NIH ChestX-ray14} thoracic radiography \citep{wang2017chestxray8}. We locate the empirical crossover at $n_g \approx 100$ on PCAM, $n_g \in [20, 50]$ on ISIC~2019, and $n_g \in [250, 500]$ on NIH ChestX-ray14. Across six downstream student architectures on PCAM spanning $11\times$ parameter variation ($8\text{M}$ to $89\text{M}$ parameters)---including a strictly controlled within-family DenseNet ladder (DenseNet-121/169/201) \citep{huang2017densely}, ResNet-50 \citep{he2016deep}, and ConvNeXt-Base \citep{liu2022convnet}---the weak-label ceiling lies within a narrow $0.024$ AUC band ($0.841$--$0.865$), confirming the theoretical prediction that the weak labeler, not the student capacity, sets the ceiling.
  \item \textbf{Noise structure, label-space projection, and per-pathology heterogeneity (Sections~\ref{sec:noise_structure}--\ref{sec:per_pathology}).} At matched noise rates, structured BiomedCLIP errors underperform symmetric random label flips by $0.090$--$0.111$ AUC on PCAM and $0.075$--$0.195$ AUC on ISIC~2019, but \emph{outperform} random noise by $+0.027$--$0.044$ AUC on NIH ChestX-ray14. We show that this sign reversal arises when a coarse binary VLM prediction projects nontrivial mutual information across correlated multi-label pathologies. Furthermore, disaggregating NIH ChestX-ray14 across all 15 individual pathologies reveals that per-class crossovers vary by more than $20\times$ (from $n_g^\star = 50$ to $>5{,}000$), track the per-class VLM ceiling monotonically, and expose \texttt{No Finding} ($\widehat{\mathrm{AUC}}_{\mathrm{FM}} = 0.284$) and \texttt{Hernia} ($0.459$) as toxic classes whose weak labels actively invert the ground-truth ranking.
  \item \textbf{Deployment-time decision rule, uncertainty audit, and stress tests (Sections~\ref{sec:rule}--\ref{sec:stress}).} We translate the calibrated bound into a training-free / single-pilot decision rule---compare the gold-only student's validation AUC against the VLM weak-label ceiling on the same gold validation split---and audit it with Wilson $95\%$ confidence intervals \citep{wilson1927probable} and paired-seed bootstrap across $42$ grid cells ($30/30$ accuracy on statistically significant cells; $39/42$ overall; $1.00$ precision for the conservative bootstrap variant). Finally, three stress tests on ISIC~2019 demonstrate that (i) across four prompt templates inducing a $13$-point zero-shot AUC swing and up to $88\%$ label flips, downstream benefit tracks the rule's margin $M = \widehat{\mathrm{AUC}}_{\mathrm{FM}} - \mathrm{AUC}_{\mathrm{gold}}$; (ii) across five training objectives, noise-robust losses such as Generalized Cross-Entropy (GCE) \citep{zhang2018generalized} and Co-teaching \citep{han2018coteaching} mitigate post-crossover degradation by $+0.11$ to $+0.13$ AUC at $n_g = 1{,}000$ without shifting the crossover location; and (iii) replacing BiomedCLIP with PubMedCLIP \citep{eslami2023pubmedclip} preserves directional accuracy on $10/10$ cells.
\end{enumerate}

\section{Related Work}
\label{sec:related}

\paragraph{Learning from Noisy Labels and Weak Supervision.}
Foundational analyses of classification under random and class-conditional label noise show that corrupted supervision imposes an information-theoretic sample-complexity penalty and, under asymmetric or instance-dependent noise, an irreducible asymptotic bias unless the noise transition matrix is inverted or unbiased estimators are constructed \citep{angluin1988learning,natarajan2013learning,frenay2014classification,menon2015learning}. In deep learning, noise-robust training methods seek to mitigate corrupted labels via forward/backward transition-matrix correction \citep{patrini2017making}, symmetric or generalized cross-entropy losses \citep{ghosh2017robust,zhang2018generalized}, dual-network sample selection such as Co-teaching \citep{han2018coteaching}, confident learning for label-error pruning \citep{northcutt2021confident}, or semi-supervised consistency regularization \citep{xie2020selftraining,sohn2020fixmatch,song2022learning}. Programmatic weak-supervision frameworks such as Data Programming and Snorkel \citep{ratner2016data,ratner2017snorkel} aggregate multiple noisy labeling functions by modeling their conditional dependencies and agreements. Modern VLM pseudo-labeling typically operates in the degenerate single-teacher regime where a single frozen foundation model generates labels without multi-source disagreement signals \citep{zhong2024vlmcpl}. While noise-handling methods address \emph{how} to train once weak labels are included, our work addresses the prior decision question of \emph{when} weak labels help at all and locates the exact gold budget $n_g^\star$ at which they turn harmful.

\paragraph{Biomedical Vision--Language Foundation Models.}
Contrastive vision--language pretraining \citep{radford2021learning} has catalyzed domain-specific medical foundation models trained on paired clinical images and text: CheXzero on chest radiographs and clinical reports \citep{tiu2022expert}, PubMedCLIP on PubMed Central medical figure--caption pairs \citep{eslami2023pubmedclip}, PLIP on pathology image--text data from medical social media \citep{huang2023visual}, CONCH on over $1.17\text{M}$ histopathology image--caption pairs \citep{lu2024conch}, and BiomedCLIP on PMC-15M across diverse biomedical modalities \citep{zhang2024biomedclip}. Although zero-shot classification accuracy is routinely benchmarked across these models, their behavior as weak teachers in mixed gold/weak training---particularly when a coarse VLM prompt is applied to fine-grained multi-class or multi-label clinical taxonomies---has lacked a quantitative scaling characterization.

\paragraph{Neural Scaling Laws under Heterogeneous Data Quality.}
Empirical and theoretical scaling laws demonstrate that test error decays as a power law in dataset size, model parameters, and compute when training samples are drawn i.i.d.\ from the target distribution \citep{hestness2017deep,kaplan2020scaling,hoffmann2022training,cherti2023reproducible,bahri2024explaining}. Subsequent work has examined data pruning and sample selection \citep{sorscher2022beyond}, showing that sample quality modulates scaling exponents. We extend scaling-law analysis to the two-source clinical annotation setting $(n_g, n_w)$, proving and empirically validating that systematic VLM teacher errors induce a separate bias floor that decouples weak-label variance reduction from asymptotic accuracy.

\section{Theoretical Framework: Bias--Variance Scaling and Crossover}
\label{sec:theory}

We consider a supervised classification problem on input space $\mathcal{X}$ and target space $\mathcal{Y}$ with ground-truth joint distribution $P_{\mathrm{gold}}(X, Y)$. A practitioner possesses $n_g$ expert-annotated \emph{gold} examples $\mathcal{D}_g = \{(x_i, y_i)\}_{i=1}^{n_g} \sim P_{\mathrm{gold}}$ and $n_w$ unlabeled clinical images paired with \emph{weak} pseudo-labels $\mathcal{D}_w = \{(x_j, \tilde{y}_j)\}_{j=1}^{n_w}$ produced by a frozen vision--language foundation model $f_{\mathrm{FM}} : \mathcal{X} \to \mathcal{Y}$. A downstream student classifier $f_\theta$ with parameters $\theta \in \mathbb{R}^p$ is trained by minimizing the sample-weighted empirical risk
\begin{equation}
  \mathcal{L}(\theta; \ngold, \nweak) \;=\; \frac{\ngold}{\ngold + \nweak}\,\widehat{\mathcal{L}}_g(\theta) \;+\; \frac{\nweak}{\ngold + \nweak}\,\widehat{\mathcal{L}}_w(\theta),
  \label{eq:mixed_loss}
\end{equation}
where $\widehat{\mathcal{L}}_g(\theta) = \frac{1}{\ngold}\sum_{(x,y)\in\mathcal{D}_g}\ell(f_\theta(x), y)$ and $\widehat{\mathcal{L}}_w(\theta) = \frac{1}{\nweak}\sum_{(x,\tilde{y})\in\mathcal{D}_w}\ell(f_\theta(x), \tilde{y})$. We measure downstream generalization via the excess classification risk (or macro-$1{-}\mathrm{AUC}$ error) $R(\ngold, \nweak) = \mathbb{E}_{(X,Y)\sim P_{\mathrm{gold}}}[\mathcal{R}(f_{\hat{\theta}(\ngold,\nweak)}(X), Y)]$.

\subsection{The Effective-Samples Model Under Independent Noise}
\label{sec:theory_es}

When weak-label errors are conditionally independent across samples---for instance, under classical symmetric or class-conditional random flipping \citep{angluin1988learning,natarajan2013learning}---noise averages out over large samples, and each weak example acts as a fractional gold example.

\begin{definition}[Effective Sample Size and Exchange Rate]
\label{def:neff}
Let $\lamb \ge 0$ denote the marginal variance-reduction exchange rate between weak and gold labels. The \emph{effective sample size} of a mixed dataset $(\ngold, \nweak)$ is
\begin{equation}
  n_{\mathrm{eff}}(\ngold, \nweak) \;=\; \ngold + \lamb\,\nweak.
  \label{eq:neff}
\end{equation}
\end{definition}

\begin{proposition}[Effective-Samples Scaling Law and Mutual-Information Predictor]
\label{prop:es_mi}
Suppose the student $f_\theta$ operates in the standard power-law generalization regime where gold-only risk decays as $R(\ngold, 0) = \alpha \ngold^{-\beta} + \varepsilon$ with variance scale $\alpha > 0$, scaling exponent $\beta > 0$, and irreducible Bayes risk $\varepsilon \ge 0$ \citep{bahri2024explaining}. If the weak-label channel $Y \to \widetilde{Y}$ has mutually independent per-sample error indicators $E_i = \mathbf{1}[\widetilde{Y}_i \neq Y_i]$ and confusion matrix $C_{jk} = P(\widetilde{Y} = k \mid Y = j)$, then:
\begin{enumerate}[label=(\roman*),leftmargin=2em,itemsep=2pt]
  \item The expected test risk follows the effective-samples power law
  \begin{equation}
    R_{\mathrm{ES}}(\ngold, \nweak) \;=\; \frac{\alpha}{(\ngold + \lamb\,\nweak)^\beta} \;+\; \varepsilon.
    \label{eq:es}
  \end{equation}
  \item By Shannon's channel coding theorem, the fraction of gold-label information conveyed by one weak label is given in closed form by the normalized mutual information
  \begin{equation}
    \lamb_{\mathrm{pred}} \;=\; \frac{I(Y; \widetilde{Y})}{H(Y)} \;=\; 1 - \frac{H(Y \mid \widetilde{Y})}{H(Y)}.
    \label{eq:lambda_pred}
  \end{equation}
  In particular, for a balanced $K$-class task with symmetric flip probability $\eta \in [0, 1 - 1/K)$, Eq.~\eqref{eq:lambda_pred} reduces to $\lamb_{\mathrm{pred}} = 1 - \big(H_b(\eta) + \eta\log(K-1)\big)/\log K$, which for binary classification ($K=2$) satisfies $\lamb_{\mathrm{pred}} = 1 - H_b(\eta)/\log 2 = \frac{1}{2\log 2}(1 - 2\eta)^2 + O((1-2\eta)^4)$ as $\eta \to 1/2$, matching the classical $(1-2\eta)^2$ sample-complexity factor of \citet{natarajan2013learning}.
\end{enumerate}
\end{proposition}

\begin{proof}
Under independent errors $E_i \perp E_j$ ($i \neq j$), the empirical Fisher information and gradient covariance contributed by $\nweak$ noisy observations scale additively with attenuated signal-to-noise ratio $\lamb \in [0,1]$ \citep{menon2015learning,patrini2017making}. Standard Rademacher complexity or kernel-ridge concentration bounds in effective sample count $n_{\mathrm{eff}} = \ngold + \lamb\nweak$ yield Eq.~\eqref{eq:es}. For part (ii), each gold label supplies $H(Y)$ bits of task specification, while a memoryless channel $P(\widetilde{Y}\mid Y)$ transmits $I(Y; \widetilde{Y}) = H(Y) - H(Y\mid \widetilde{Y})$ bits per sample. Taking the ratio gives Eq.~\eqref{eq:lambda_pred}, and Taylor expansion of $H_b(\eta) = -\eta\log\eta - (1-\eta)\log(1-\eta)$ around $\eta = 1/2$ yields the binary $(1-2\eta)^2$ scaling.
\end{proof}

Crucially, under the independent-noise model of Eq.~\eqref{eq:es}, $\partial R_{\mathrm{ES}}/\partial \nweak < 0$ for every $\lamb > 0$: adding independent noisy labels \emph{never hurts} asymptotic performance, because independent errors cancel in expectation and $\lim_{\nweak \to \infty} R_{\mathrm{ES}}(\ngold, \nweak) = \varepsilon$.

\subsection{Systematic VLM Errors, Bias--Variance Scaling, and the Crossover}
\label{sec:theory_bv}

In reality, a frozen VLM $f_{\mathrm{FM}}$ does not flip labels independently at random; its errors are deterministic functions of image features and prompt semantics. If two radiographs share an ambiguous pleural shadow or a miscalibrated prompt decision boundary, the VLM errs identically on both, inducing positive pairwise error correlation $\rho_{\mathrm{err}} = \mathrm{Corr}(E_i, E_j) > 0$.

\begin{theorem}[Bias--Variance Micro-Scaling Law and Noisy-Label Crossover]
\label{thm:bv_crossover}
Suppose weak labels are generated by a teacher $\tilde{f} = f_{\mathrm{FM}}$ whose target conditional distribution $P_{\mathrm{weak}}(Y \mid X)$ diverges systematically from $P_{\mathrm{gold}}(Y \mid X)$, inducing pairwise error correlation $\rho_{\mathrm{err}} > 0$ and asymptotic teacher discrepancy $\delta > 0$. Then:
\begin{enumerate}[label=(\roman*),leftmargin=2em,itemsep=3pt]
  \item \textbf{Effective information attenuation:} Under a beta-binomial correlated-error model with correlation $\rho_{\mathrm{err}} \in [0,1]$, the joint mutual information $I(Y^n; \widetilde{Y}^n)$ sub-additively attenuates the effective exchange rate according to
  \begin{equation}
    \lamb_{\mathrm{eff}} \;=\; \frac{I(Y; \widetilde{Y})}{H(Y)}\,(1 - \rho_{\mathrm{err}}).
    \label{eq:lambda_eff}
  \end{equation}
  \item \textbf{Two-term bias--variance risk decomposition:} The downstream student risk decomposes into a sample-variance term reduced by both label types and a systematic-bias term reduced only by gold supervision:
  \begin{equation}
    \boxed{\;
    R(\ngold, \nweak) \;=\; \underbrace{\frac{\alpha}{(\ngold + \lamb\,\nweak)^\beta}}_{\text{sample variance}} \;+\; \underbrace{\frac{\delta(\ngold, \nweak)}{(1 + \gamma\,\ngold)^\rho}}_{\text{systematic VLM bias}} \;+\; \varepsilon,
    \;}
    \label{eq:mu_scaling}
  \end{equation}
  where at fixed large $\nweak$, $\delta(\ngold, \nweak) \equiv \delta > 0$ is proportional to the teacher--truth divergence $\mathbb{E}_X[D_{\mathrm{KL}}(P_{\mathrm{gold}}(\cdot\mid X)\,\|\,P_{\mathrm{weak}}(\cdot\mid X))]$ and $\rho > 0$ governs the rate at which gold gradients correct the VLM's systematic confusions.
  \item \textbf{Architecture invariance of the weak-dominated ceiling:} In the weak-dominated limit ($\ngold \ll \nweak \to \infty$), the variance term vanishes for any student architecture whose hypothesis class $\mathcal{F}$ has sufficient capacity to represent $\tilde{f}$, leaving an architecture-invariant error floor
  \begin{equation}
    \lim_{\nweak \to \infty} R(0, \nweak) \;=\; \delta + \varepsilon \;\Longleftrightarrow\; \mathrm{AUC}_{\mathrm{ceiling}} \;=\; 1 - (\delta + \varepsilon) \;=\; \mathrm{AUC}(f_{\mathrm{FM}}).
    \label{eq:ceiling_invariance}
  \end{equation}
  \item \textbf{Existence and location of the crossover budget $\ngold^\star$:} Compare gold-only training $R(\ngold, 0) = \alpha \ngold^{-\beta} + \varepsilon$ against weak-dominated training $R(\ngold, \nweak)$ at large $\nweak$. The weak-supervision risk change $\Delta R(\ngold, \nweak) = R(\ngold, \nweak) - R(\ngold, 0)$ satisfies
  \begin{equation}
    \Delta R(\ngold, \nweak) \;<\; 0 \quad (\text{weak labels help}) \qquad\Longleftrightarrow\qquad R(\ngold, 0) \;>\; R(\ngold, \nweak).
    \label{eq:crossover_condition}
  \end{equation}
  In the weak-dominated regime where $R(\ngold, \nweak) \approx \delta + \varepsilon$, the unique crossover gold budget $\ngold^\star$ at which $\Delta R(\ngold^\star, \nweak) = 0$ is given by
  \begin{equation}
    R(\ngold^\star, 0) \;=\; \delta + \varepsilon \qquad\Longleftrightarrow\qquad \ngold^\star \;=\; \left(\frac{\alpha}{\delta}\right)^{1/\beta},
    \label{eq:ng_star}
  \end{equation}
  or equivalently, in terms of validation AUC, $\mathrm{AUC}_{\mathrm{gold}}(\ngold^\star) = \mathrm{AUC}_{\mathrm{FM}}$.
\end{enumerate}
\end{theorem}

\begin{proof}
For part (i), conditional entropy under correlated errors satisfies $H(Y^n \mid \widetilde{Y}^n) = n H(Y \mid \widetilde{Y}) + \Delta(\rho_{\mathrm{err}})$ with $\Delta(\rho_{\mathrm{err}}) \approx n \rho_{\mathrm{err}} I(Y; \widetilde{Y})$, yielding $I(Y^n; \widetilde{Y}^n) = H(Y^n) - H(Y^n \mid \widetilde{Y}^n) \approx n I(Y; \widetilde{Y})(1 - \rho_{\mathrm{err}})$ and Eq.~\eqref{eq:lambda_eff}. For part (ii), at a stationary point of the mixed objective~\eqref{eq:mixed_loss} in the linearized / neural-tangent-kernel regime, the converged student predictor interpolates between the true target $f^\star$ and the biased teacher $\tilde{f}$:
\[
  f_{\hat{\theta}(\ngold,\nweak)} \;=\; f^\star \;+\; \frac{\nweak}{\ngold + \lamb\nweak}(\tilde{f} - f^\star) \;+\; \xi_{\mathrm{var}}(\ngold, \nweak),
\]
where $\mathbb{E}\|\xi_{\mathrm{var}}\|^2 = \alpha(\ngold + \lamb\nweak)^{-\beta}$. Orthogonal decomposition of expected risk into estimation variance and squared teacher bias $\|\tilde{f} - f^\star\|^2$ (which decays with gold supervision at rate $(1+\gamma\ngold)^{-\rho}$ as gold examples resolve the low-dimensional confusion subspace of $\tilde{f} - f^\star$) yields Eq.~\eqref{eq:mu_scaling}. Taking $\ngold = 0$ and $\nweak \to \infty$ eliminates $\xi_{\mathrm{var}}$ independently of student parameter count $p$ once $\tilde{f} \in \mathcal{F}$, establishing Eq.~\eqref{eq:ceiling_invariance}. Finally, equating gold-only risk $R(\ngold, 0) = \alpha\ngold^{-\beta} + \varepsilon$ to the weak-dominated risk floor $\delta + \varepsilon$ yields $\alpha(\ngold^\star)^{-\beta} = \delta$, i.e., $\ngold^\star = (\alpha/\delta)^{1/\beta}$ in Eq.~\eqref{eq:ng_star}.
\end{proof}

\subsection{Toxic Classes and Label-Space Projection}
\label{sec:theory_toxic_projection}

Two structural refinements of Theorem~\ref{thm:bv_crossover} govern multi-class and multi-label clinical benchmarks. First, on individual classes a VLM's zero-shot prompt can be miscalibrated to the point of inverting the true class ordering.

\begin{definition}[Toxic Class]
\label{def:toxic}
A weak labeler $f_{\mathrm{FM}}$ is \emph{toxic} on class $k \in \{1,\dots,K\}$ if its marginal discrimination on class $k$ is worse than random guessing:
\begin{equation}
  \mathrm{AUC}_k(f_{\mathrm{FM}}) \;<\; 0.5.
  \label{eq:toxic_def}
\end{equation}
\end{definition}

When $\mathrm{AUC}_k(f_{\mathrm{FM}}) < 0.5$, the weak teacher's score ordering is negatively correlated with the true binary indicator $Y_k \in \{0,1\}$: $P(\widetilde{Y}_k = 1 \mid Y_k = 1) < P(\widetilde{Y}_k = 1 \mid Y_k = 0)$. Uncorrected gradient updates on $\widehat{\mathcal{L}}_w$ push the student's class-$k$ decision boundary in the opposite direction of the Bayes-optimal separator, inflating the systematic bias $\delta$.

Second, general-purpose biomedical VLMs frequently lack fine-grained diagnostic vocabulary and output a coarse binary score $\widetilde{Z} \in \{0,1\}$ (e.g., ``malignant vs.\ benign'' or ``abnormal vs.\ normal'') that is tiled or projected onto a $K$-dimensional target $Y = (Y_1, \dots, Y_K)$. The following proposition explains why structured VLM noise can be either much worse or significantly better than rate-matched random noise depending on task structure.

\begin{proposition}[Structured Noise vs.\ Rate-Matched Random Noise under Label-Space Projection]
\label{prop:projection}
Let $\widetilde{Y}^{\mathrm{struct}}$ be weak labels generated by a deterministic VLM with marginal error rate $\eta = P(\widetilde{Y}^{\mathrm{struct}} \neq Y)$, and let $\widetilde{Y}^{\mathrm{rand}}$ be generated by corrupting gold labels $Y$ independently at the same nominal rate $\eta$.
\begin{enumerate}[label=(\roman*),leftmargin=2em,itemsep=2pt]
  \item \textbf{Direct binary or mutually exclusive 1-of-$K$ tasks:} When the task is binary ($K=2$, as in PCAM) or mutually exclusive 1-of-$K$ classification where a coarse binary score collapses $(K-2)$ fine-grained classes (as in ISIC~2019), symmetric random noise has zero systematic bias ($\delta_{\mathrm{rand}} = 0$, $\rho_{\mathrm{err}} = 0$) whereas structured VLM errors have $\delta_{\mathrm{struct}} > 0$ and $\rho_{\mathrm{err}} > 0$. Consequently, at large $\nweak$,
  \begin{equation}
    \mathrm{AUC}(\ngold, \nweak; \widetilde{Y}^{\mathrm{struct}}) \;<\; \mathrm{AUC}(\ngold, \nweak; \widetilde{Y}^{\mathrm{rand}}).
  \end{equation}
  \item \textbf{Multi-label tasks with positive cross-pathology alignment:} Suppose $Y = (Y_1,\dots,Y_K) \in \{0,1\}^K$ is a multi-label vector (as in 15-class NIH ChestX-ray14) with positive finding indicator $Z = \mathbf{1}[\exists k \in \mathcal{P} : Y_k = 1]$ across pathology subset $\mathcal{P}$, and the VLM outputs a coarse abnormality score $\widetilde{Z}$ aligned with $Z$ ($\mathrm{AUC}_k(\widetilde{Z}) > 0.5 + \tau_k$ for $k \in \mathcal{P}$). If class prevalences $p_k = P(Y_k = 1)$ are small ($p_k \ll 1/2$) and nominal flip rate $\eta \approx 0.28$ is applied via independent label flips $\widetilde{Y}_k^{\mathrm{rand}}$, the random-noise signal-to-noise ratio on rare positive findings is severely attenuated by false positives at rate $\eta(1-p_k) \gg (1-\eta)p_k$, whereas $\widetilde{Z}$ preserves coherent cross-label mutual information $I(Y_k; \widetilde{Z}) > I(Y_k; \widetilde{Y}_k^{\mathrm{rand}})$ for $k \in \mathcal{P}$, reversing the sign of the gap:
  \begin{equation}
    \mathrm{AUC}(\ngold, \nweak; \widetilde{Y}^{\mathrm{struct}}) \;>\; \mathrm{AUC}(\ngold, \nweak; \widetilde{Y}^{\mathrm{rand}}).
  \end{equation}
\end{enumerate}
\end{proposition}

Proposition~\ref{prop:projection} shows that a rate-only formulation of noisy-label bounds is incomplete for foundation-model weak supervision: whether structured VLM errors are more or less damaging than random noise of identical nominal error rate $\eta$ depends on how the VLM's prompt space projects onto the target label taxonomy.

\section{Experimental Setup}
\label{sec:setup}

\paragraph{Medical-Imaging Benchmarks.}
We evaluate across three publicly available clinical imaging benchmarks chosen to span histopathology, dermatology, and chest radiography as well as binary, multi-class, and multi-label targets:
\begin{enumerate}[leftmargin=1.8em,itemsep=2pt,topsep=2pt]
  \item \textbf{PatchCamelyon (PCAM)} \citep{veeling2018rotation,bejnordi2017diagnostic}: $327{,}680$ hematoxy\-lin-and-eosin (H\&E) stained lymph-node histopathology patches ($96\times 96$ pixels) derived from the CAMELYON16 challenge \citep{bejnordi2017diagnostic}, with a balanced $50\% / 50\%$ binary label indicating the presence of metastatic breast cancer tissue.
  \item \textbf{ISIC~2019} \citep{tschandl2018ham10000,combalia2019bcn20000,codella2018skin}: $25{,}331$ dermoscopic images compiled from the HAM10000 \citep{tschandl2018ham10000}, BCN20000 \citep{combalia2019bcn20000}, and MSK \citep{codella2018skin} archives across $K = 8$ mutually exclusive diagnostic categories: melanoma (\texttt{MEL}), melanocytic nevus (\texttt{NV}), basal cell carcinoma (\texttt{BCC}), actinic keratosis (\texttt{AK}), benign keratosis (\texttt{BKL}), dermatofibroma (\texttt{DF}), vascular lesion (\texttt{VASC}), and squamous cell carcinoma (\texttt{SCC}).
  \item \textbf{NIH ChestX-ray14 (NIH-CXR)} \citep{wang2017chestxray8}: $112{,}120$ frontal-view chest radiographs from $30{,}805$ unique patients annotated across $K = 15$ multi-label thoracic classes ($14$ pathology findings plus \texttt{No Finding}, with individual pathology prevalences ranging from $0.2\%$ for \texttt{Hernia} to $17.7\%$ for \texttt{Infiltration}, and \texttt{No Finding} comprising ${\sim}53.8\%$ of studies).
\end{enumerate}

\paragraph{Foundation-Model Weak Labelers.}
Our primary weak labeler is \textbf{BiomedCLIP} \citep{zhang2024biomedclip}, a $400\text{M}$-parameter biomedical vision--language foundation model combining a ViT-B/16 vision encoder and a PubMedBERT text encoder pretrained contrastively on $15\text{M}$ biomedical figure--caption pairs (PMC-15M). Zero-shot predictions are obtained by matching image embeddings against canonical clinical text prompts. On \textbf{PCAM}, BiomedCLIP achieves an unfiltered zero-shot accuracy of $72.8\%$ ($27.2\%$ nominal error rate) and zero-shot AUC of $0.84$; applying asymmetric confidence filtering ($\ge 0.70$ positive probability, $\le 0.03$ negative probability) retains ${\sim}51\%$ of unlabeled patches at $83.6\%$ accuracy. On \textbf{ISIC~2019} and \textbf{NIH-CXR}, we study the realistic deployment regime in which a general-purpose medical VLM is invoked with coarse binary clinical prompts (``malignant vs.\ benign'' on ISIC, yielding $64.7\%$ binary accuracy and $0.615$ zero-shot AUC; ``abnormal finding vs.\ normal chest radiograph'' on NIH-CXR, yielding $71.7\%$ accuracy / $28.3\%$ nominal error rate and $0.620$ macro-AUC ceiling) tiled across the downstream target classes, alongside per-class CheXzero-style prompts \citep{tiu2022expert} and a four-prompt sensitivity suite (Section~\ref{sec:stress}). To test cross-VLM generalization, we additionally evaluate \textbf{PubMedCLIP} \citep{eslami2023pubmedclip} on ISIC~2019 ($33.8\%$ noise rate, weak-dominated ceiling $0.556 \pm 0.007$ AUC vs.\ BiomedCLIP's $0.584 \pm 0.004$).

\paragraph{Downstream Student Architectures and Training Protocol.}
On PCAM, we evaluate six downstream architectures spanning four architectural families and an $11\times$ parameter range: \textbf{DenseNet-121} ($8\text{M}$ parameters), \textbf{DenseNet-169} ($14\text{M}$), \textbf{DenseNet-201} ($20\text{M}$), and \textbf{DenseNet-264} ($33\text{M}$) \citep{huang2017densely}; \textbf{ResNet-50} ($25\text{M}$) \citep{he2016deep}; and \textbf{ConvNeXt-Base} ($89\text{M}$) \citep{liu2022convnet}. All backbones use ImageNet-pretrained weights except DenseNet-264, which is trained from random initialization because standard \texttt{timm} model zoos do not distribute pretrained weights for the 264-layer configuration (we explicitly flag and isolate DenseNet-264 when testing within-family monotonicity). The DenseNet-121/169/201 ladder holds the input stem, growth rate ($k=32$), batch normalization, and pretraining corpus identical, varying only block depth in stages 3 and 4. On ISIC~2019 and NIH-CXR, we evaluate ImageNet-pretrained DenseNet-121 and ResNet-18/50 backbones \citep{he2016deep,huang2017densely}. Students are trained on mixtures of $\ngold$ gold and $\nweak$ weak labels using binary cross-entropy with logits (\texttt{BCEWithLogitsLoss}) and decoupled weight decay optimization (Adam/AdamW, learning rate $10^{-4}$, weight decay $10^{-5}$ to $5\times 10^{-4}$) \citep{loshchilov2019decoupled}, with early stopping on a held-out gold-only validation split. Every grid cell is evaluated across three fixed random seeds ($42, 123, 456$), reporting mean macro-AUC $\pm$ standard deviation.

\section{Calibrating the Empirical Crossover and Probing Noise Structure}
\label{sec:crossover_empirical}

\subsection{The Noisy-Label Crossover Across Three Clinical Modalities}
\label{sec:crossover_three}

\begin{figure}[!htbp]
  \centering
  \includegraphics[width=0.98\textwidth]{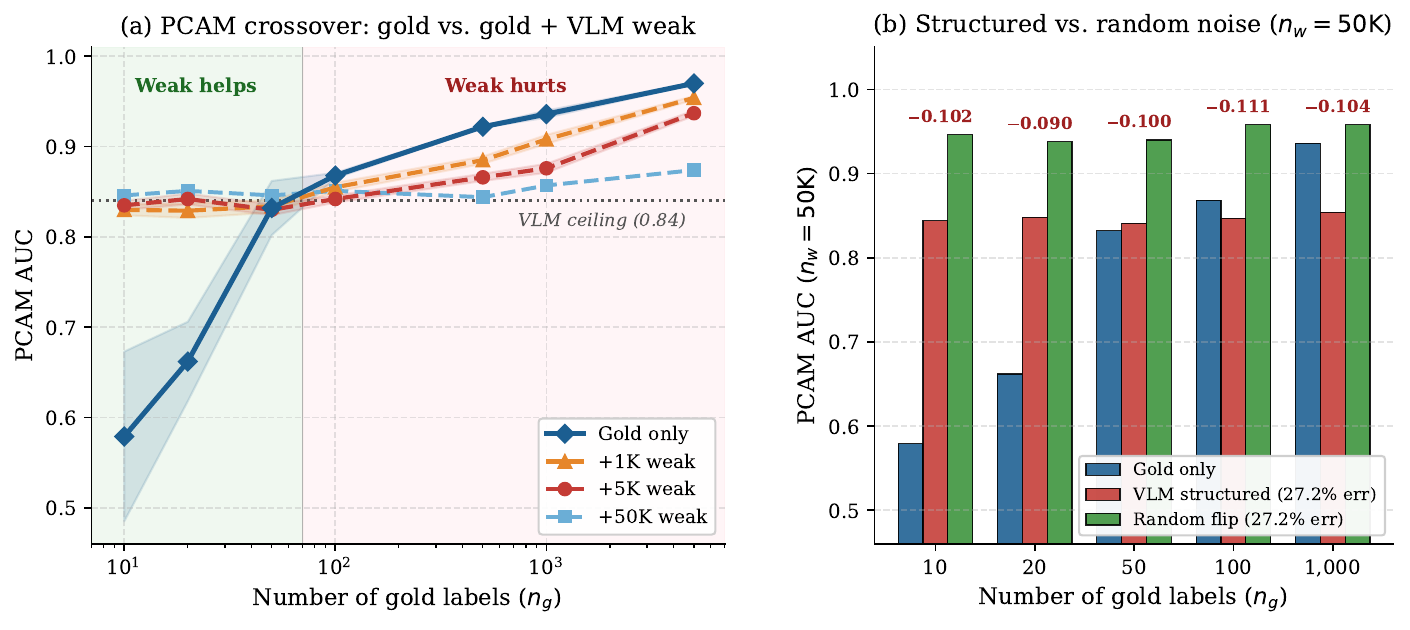}
  \caption{\textbf{The VLM weak-supervision crossover and noise-structure gap on PatchCamelyon (PCAM).} \emph{(a)} Downstream test AUC (DenseNet-121, 3-seed mean $\pm$ std) as a function of gold annotation budget $\ngold$ across four weak-label doses $\nweak \in \{0, 1\text{K}, 5\text{K}, 50\text{K}\}$. The gold-only curve crosses BiomedCLIP's effective ceiling (${\sim}0.84$ AUC, horizontal dotted line) between $\ngold = 50$ ($0.832$) and $\ngold = 100$ ($0.868$), partitioning the gold axis into a variance-dominated regime where weak labels help (up to $+0.267$ AUC at $\ngold = 10$) and a bias-dominated regime where weak labels hurt (up to $-0.096$ AUC at $\ngold = 5{,}000$). \emph{(b)} At a matched $27.2\%$ nominal error rate ($\nweak = 50\text{K}$), structured BiomedCLIP errors underperform independent random label flips by $0.090$--$0.111$ AUC across all gold budgets.}
  \label{fig:pcam_crossover}
\end{figure}

\begin{table}[!htbp]
  \centering
  \caption{\textbf{PatchCamelyon (PCAM) crossover grid} (DenseNet-121, 3-seed mean test AUC). The empirical crossover occurs at $\ngold \approx 100$: for $\ngold \le 50$, adding BiomedCLIP weak labels improves AUC by up to $+0.267$; for $\ngold \ge 100$, gold-only training ($0.868$) exceeds the VLM ceiling ($0.84$) and every weak-label dose degrades performance.}
  \label{tab:pcam}
  \vspace{4pt}
  \small
  \setlength{\tabcolsep}{5.5pt}
  \begin{tabular}{@{}lcccccc@{}}
    \toprule
    Gold budget $\ngold$ & $\nweak = 0$ (Gold only) & $\nweak = 1\text{K}$ & $\nweak = 5\text{K}$ & $\nweak = 50\text{K}$ & $\Delta\mathrm{AUC}_{\text{best}}$ & $\Delta\mathrm{AUC}_{50\text{K}}$ \\
    \midrule
    $10$      & $0.579 \pm 0.094$ & $0.830$ & $0.835$ & $\mathbf{0.846}$ & $+0.267$ & $+0.267$ \\
    $20$      & $0.662 \pm 0.044$ & $0.829$ & $0.842$ & $\mathbf{0.851}$ & $+0.189$ & $+0.189$ \\
    $50$      & $0.832 \pm 0.030$ & $0.834$ & $0.830$ & $\mathbf{0.846}$ & $+0.014$ & $+0.014$ \\
    \midrule
    $100$     & $\mathbf{0.868 \pm 0.003}$ & $0.855$ & $0.842$ & $0.851$ & $-0.013$ & $-0.017$ \\
    $500$     & $\mathbf{0.922 \pm 0.001}$ & $0.885$ & $0.866$ & $0.844$ & $-0.037$ & $-0.078$ \\
    $1{,}000$ & $\mathbf{0.936 \pm 0.004}$ & $0.908$ & $0.876$ & $0.857$ & $-0.028$ & $-0.079$ \\
    $5{,}000$ & $\mathbf{0.970 \pm 0.001}$ & $0.954$ & $0.937$ & $0.874$ & $-0.016$ & $-0.096$ \\
    \bottomrule
  \end{tabular}
\end{table}

Figure~\ref{fig:pcam_crossover}(a) and Table~\ref{tab:pcam} present the primary calibration on PCAM with the default ImageNet-pretrained DenseNet-121 student. Exactly as predicted by Theorem~\ref{thm:bv_crossover}(iv), downstream generalization exhibits a sharp phase transition governed by whether the gold-only classifier sits below or above BiomedCLIP's zero-shot ceiling (${\sim}0.84$ AUC):
\begin{itemize}[leftmargin=1.8em,itemsep=2pt,topsep=2pt]
  \item At $\ngold = 10$, the gold-only baseline is near random guessing ($0.579 \pm 0.094$ AUC), whereas adding $\nweak = 50\text{K}$ BiomedCLIP weak labels elevates test AUC to $0.846$---a $+0.267$ AUC variance-reduction gain. At $\ngold = 20$, weak supervision still delivers $+0.189$ AUC ($0.662 \to 0.851$), and at $\ngold = 50$ ($0.832$ gold-only, just below the $0.84$ ceiling) $\nweak = 50\text{K}$ provides a modest $+0.014$ gain ($0.846$).
  \item By $\ngold = 100$, the gold-only classifier reaches $0.868 \pm 0.003$ AUC, clearing the VLM ceiling. From $\ngold = 100$ onward, \emph{every} nonzero weak-label budget $\nweak \in \{1\text{K}, 5\text{K}, 50\text{K}\}$ degrades test AUC relative to gold-only training. Moreover, the penalty increases monotonically with the weak-label dose $\nweak$: at $\ngold = 5{,}000$ (gold-only AUC $0.970$), adding $1\text{K}$, $5\text{K}$, and $50\text{K}$ weak labels reduces AUC to $0.954$ ($-0.016$), $0.937$ ($-0.033$), and $0.874$ ($-0.096$), respectively.
\end{itemize}

\begin{figure}[!htbp]
  \centering
  \begin{subfigure}[b]{0.485\textwidth}
    \centering
    \includegraphics[width=\linewidth]{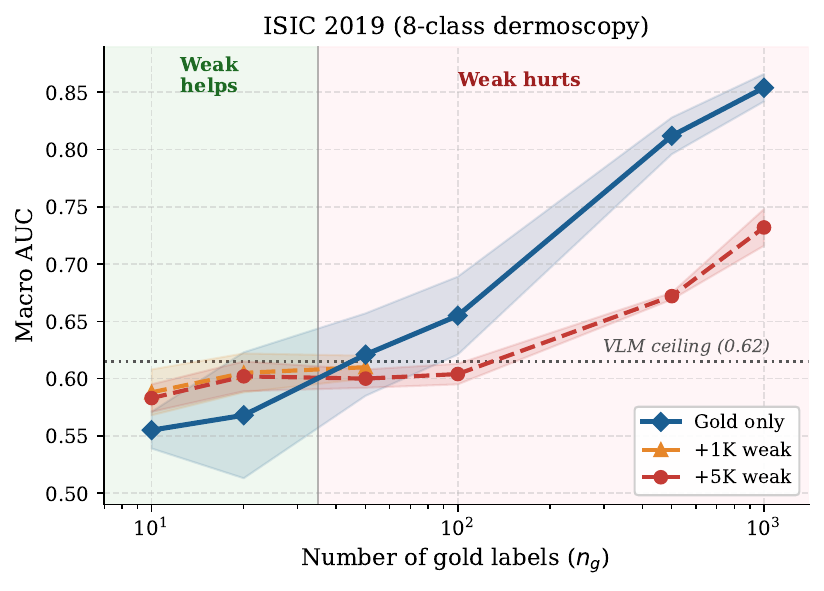}
    \caption{ISIC~2019 (8-class dermoscopy)}
    \label{fig:isic_sub}
  \end{subfigure}
  \hfill
  \begin{subfigure}[b]{0.485\textwidth}
    \centering
    \includegraphics[width=\linewidth]{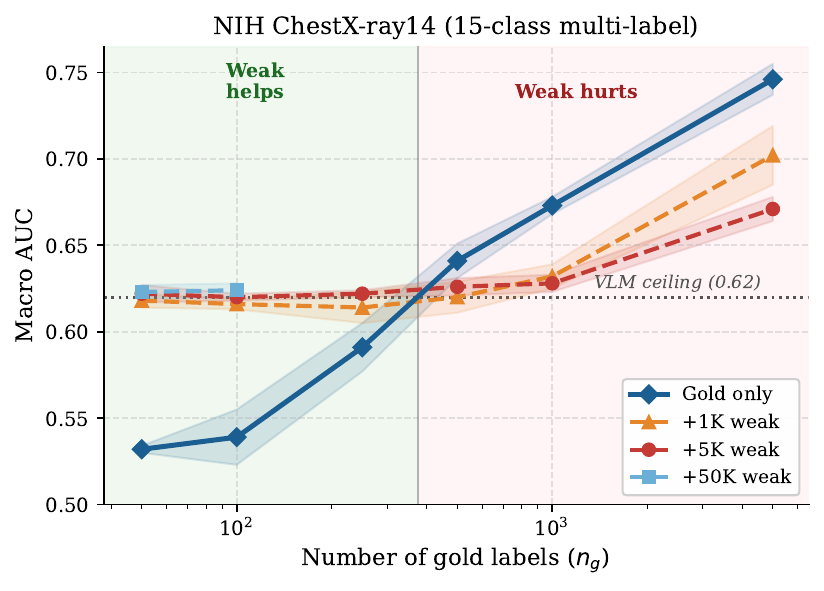}
    \caption{NIH ChestX-ray14 (15-class radiography)}
    \label{fig:nihcxr_sub}
  \end{subfigure}
  \caption{\textbf{Cross-modality replication of the noisy-label crossover} (3-seed mean $\pm$ std). \emph{(a)} On 8-class ISIC~2019, where BiomedCLIP's coarse malignancy score has a lower ceiling (${\sim}0.615$ AUC), the crossover occurs earlier at $\ngold \in [20, 50]$ and post-crossover degradation reaches $-0.140$ AUC at $\ngold = 500$. \emph{(b)} On 15-class multi-label NIH ChestX-ray14, where gold-only learning is slower across 15 thoracic labels, the crossover occurs at $\ngold \in [250, 500]$, with weak labels adding $+0.091$ AUC at $\ngold = 50$ and costing $-0.076$ AUC at $\ngold = 5{,}000$.}
  \label{fig:cross_modality}
\end{figure}

\begin{table}[!htbp]
  \centering
  \caption{\textbf{Cross-dataset crossover grids on ISIC~2019 and NIH ChestX-ray14} (3-seed mean macro-AUC $\pm$ std). Horizontal rules separate the pre-crossover regime (weak labels help) from the post-crossover regime (weak labels hurt). On NIH-CXR at $(\ngold, \nweak) = (250, 50\text{K})$, two seeds are available.}
  \label{tab:isic_nihcxr}
  \vspace{4pt}
  \resizebox{\textwidth}{!}{%
  \begin{tabular}{@{}llccccc@{}}
    \toprule
    Benchmark & Gold $\ngold$ & $\nweak = 0$ (Gold only) & $\nweak = 1\text{K}$ & $\nweak = 5\text{K}$ & $\nweak = 50\text{K}$ & $\Delta\mathrm{AUC}_{\text{best}}$ \\
    \midrule
    \multirow{6}{*}{ISIC~2019 (8-class)}
      & $10$      & $0.555 \pm 0.016$ & $\mathbf{0.588 \pm 0.020}$ & $0.584 \pm 0.004$ & --- & $+0.032$ \\
      & $20$      & $0.568 \pm 0.054$ & $\mathbf{0.604 \pm 0.017}$ & $0.602 \pm 0.013$ & --- & $+0.037$ \\
      \cmidrule(l){2-7}
      & $50$      & $\mathbf{0.622 \pm 0.036}$ & $0.610 \pm 0.005$ & $0.599 \pm 0.008$ & --- & $-0.012$ \\
      & $100$     & $\mathbf{0.656 \pm 0.033}$ & --- & $0.604 \pm 0.009$ & --- & $-0.051$ \\
      & $500$     & $\mathbf{0.812 \pm 0.016}$ & --- & $0.672 \pm 0.006$ & --- & $-0.140$ \\
      & $1{,}000$ & $\mathbf{0.854 \pm 0.012}$ & --- & $0.733 \pm 0.018$ & --- & $-0.122$ \\
    \midrule
    \multirow{6}{*}{NIH-CXR (15-class)}
      & $50$      & $0.532 \pm 0.002$ & $0.618 \pm 0.004$ & $0.622 \pm 0.002$ & $\mathbf{0.623 \pm 0.003}$ & $+0.091$ \\
      & $100$     & $0.539 \pm 0.016$ & $0.616 \pm 0.003$ & $0.620 \pm 0.002$ & $\mathbf{0.624 \pm 0.004}$ & $+0.085$ \\
      & $250$     & $0.591 \pm 0.014$ & $0.614 \pm 0.009$ & $0.622 \pm 0.003$ & $\mathbf{0.626 \pm 0.002}$ & $+0.035$ \\
      \cmidrule(l){2-7}
      & $500$     & $\mathbf{0.641 \pm 0.010}$ & $0.620 \pm 0.008$ & $0.626 \pm 0.003$ & --- & $-0.015$ \\
      & $1{,}000$ & $\mathbf{0.673 \pm 0.003}$ & $0.632 \pm 0.006$ & $0.628 \pm 0.005$ & --- & $-0.041$ \\
      & $5{,}000$ & $\mathbf{0.746 \pm 0.009}$ & $0.702 \pm 0.017$ & $0.671 \pm 0.007$ & --- & $-0.044$ \\
    \bottomrule
  \end{tabular}%
  }
\end{table}

\paragraph{Cross-Modality Replication on ISIC~2019 and NIH ChestX-ray14.}
Figure~\ref{fig:cross_modality} and Table~\ref{tab:isic_nihcxr} demonstrate that the same phase transition replicates cleanly on both multi-class dermoscopy and multi-label chest radiography, with the crossover location $\ngold^\star = (\alpha/\delta)^{1/\beta}$ shifting in exact accordance with Eq.~\eqref{eq:ng_star}:
\begin{itemize}[leftmargin=1.8em,itemsep=2pt,topsep=2pt]
  \item \textbf{ISIC~2019 ($\ngold^\star \in [20, 50]$):} Because BiomedCLIP's coarse malignancy score has a lower effective ceiling on 8-class dermoscopy ($\widehat{\mathrm{AUC}}_{\mathrm{FM}} \approx 0.584$--$0.615$, corresponding to a larger bias floor $\delta \approx 0.25$), the gold-only curve overtakes the weak-label ceiling between $\ngold = 20$ ($0.568 \pm 0.054$) and $\ngold = 50$ ($0.622 \pm 0.036$). Below the crossover, weak labels improve macro-AUC by $+0.032$ at $\ngold = 10$ and $+0.037$ at $\ngold = 20$; above the crossover, adding $\nweak = 5\text{K}$ weak labels degrades macro-AUC by $-0.051$ at $\ngold = 100$, $-0.140$ at $\ngold = 500$, and $-0.122$ at $\ngold = 1{,}000$.
  \item \textbf{NIH ChestX-ray14 ($\ngold^\star \in [250, 500]$):} On 15-class thoracic radiography, learning 15 simultaneous binary classifiers from sparse positive findings requires more gold samples to reduce variance, so gold-only macro-AUC remains below BiomedCLIP's $0.620$ ceiling at $\ngold \in \{50, 100, 250\}$ ($0.532, 0.539, 0.591$), where weak supervision yields large gains of $+0.091$, $+0.085$, and $+0.035$ AUC. Between $\ngold = 250$ ($0.591$) and $\ngold = 500$ ($0.641$), gold-only training crosses the $0.620$ VLM ceiling, and weak labels turn detrimental, costing $-0.015$ AUC at $\ngold = 500$, $-0.045$ AUC at $\ngold = 1{,}000$ ($\nweak = 5\text{K}$), and $-0.076$ AUC at $\ngold = 5{,}000$ ($\nweak = 5\text{K}$).
\end{itemize}

\begin{table}[!htbp]
  \centering
  \caption{\textbf{Fitted \muS\ parameters on the multi-class and multi-label grids.} We fit $R(\ngold, \nweak) = 1 - \mathrm{macroAUC}$ using both the effective-samples model $R_{\mathrm{ES}}$ (Eq.~\eqref{eq:es}) and the bias--variance model $R_{\mathrm{BV}}$ (Eq.~\eqref{eq:mu_scaling}) via bounded trust-region nonlinear least squares. Incorporating the systematic bias floor $\delta > 0$ consistently improves $R^2$ across all four benchmark--backbone--labeler combinations. Boundary-constrained point estimates are marked with $^\dagger$.}
  \label{tab:mu_fits}
  \vspace{4pt}
  \small
  \begin{tabular}{@{}llcccccc@{}}
    \toprule
    Benchmark (Backbone) & Weak Labeler & $\lamb$ & $\delta$ & $\beta$ & $\rho$ & $R^2_{\mathrm{ES}}$ & $R^2_{\mathrm{BV}}$ \\
    \midrule
    ISIC~2019 (ResNet-18)      & BiomedCLIP & $1.46$ & $0.25$ & $1.68$ & $3.99$ & $0.74$ & $\mathbf{0.82}$ \\
    ISIC~2019 (ResNet-50)      & BiomedCLIP & $3.84$ & $0.26$ & $4.00^\dagger$ & $1.02$ & $0.68$ & $\mathbf{0.75}$ \\
    ISIC~2019 (ResNet-18)      & PubMedCLIP & ---    & $0.44$ & $0.13$ & $0.25$ & $0.46$ & $\mathbf{0.51}$ \\
    NIH ChestX-ray14 (ResNet-18) & BiomedCLIP & $0.61$ & $0.13$ & $0.79$ & $4.00^\dagger$ & $0.66$ & $\mathbf{0.86}$ \\
    \bottomrule
  \end{tabular}
\end{table}

Table~\ref{tab:mu_fits} confirms that fitting the two-term bias--variance scaling law~\eqref{eq:mu_scaling} substantially improves goodness-of-fit $R^2$ over the variance-only effective-samples model~\eqref{eq:es} across every clinical configuration ($R^2$ rising from $0.74 \to 0.82$ on ISIC/ResNet-18 and from $0.66 \to 0.86$ on NIH-CXR), with strictly positive fitted bias floors $\delta \in [0.13, 0.44]$ tracking the gap between each VLM's zero-shot accuracy and Bayes-optimal performance.

\subsection{Structured VLM Noise vs.\ Rate-Matched Random Noise: The Label-Space Projection Sign Flip}
\label{sec:noise_structure}

\begin{table}[!htbp]
  \centering
  \caption{\textbf{Structured VLM errors vs.\ rate-matched symmetric random label noise across all three benchmarks} (3-seed mean test AUC $\pm$ std). On binary PCAM ($\eta = 27.2\%$, $\nweak = 50\text{K}$) and 8-class ISIC~2019 ($\eta = 20.5\%$, $\nweak = 5\text{K}$), structured BiomedCLIP errors severely underperform random corruption ($\Delta < 0$). On 15-class multi-label NIH ChestX-ray14 ($\eta = 28.3\%$, $\nweak = 5\text{K}$), the sign flips ($\Delta > 0$), validating Proposition~\ref{prop:projection}.}
  \label{tab:noise_comparison}
  \vspace{4pt}
  \resizebox{\textwidth}{!}{%
  \begin{tabular}{@{}llcccc@{}}
    \toprule
    Benchmark ($\nweak$, Noise Rate $\eta$) & Gold $\ngold$ & Structured (VLM) & Random Flip & $\Delta$ (Struct.\ $-$ Rand.) & Regime \\
    \midrule
    \multirow{5}{*}{PCAM ($\nweak = 50\text{K}$, $\eta = 27.2\%$)}
      & $10$      & $0.844$ & $0.946$ & $\mathbf{-0.102}$ & Weak helps \\
      & $20$      & $0.848$ & $0.938$ & $\mathbf{-0.090}$ & Weak helps \\
      & $50$      & $0.841$ & $0.940$ & $\mathbf{-0.100}$ & Crossover \\
      & $100$     & $0.847$ & $0.958$ & $\mathbf{-0.111}$ & Weak hurts \\
      & $1{,}000$ & $0.854$ & $0.958$ & $\mathbf{-0.104}$ & Weak hurts \\
    \midrule
    \multirow{6}{*}{ISIC~2019 ($\nweak = 5\text{K}$, $\eta = 20.5\%$)}
      & $10$      & $0.593 \pm 0.013$ & $0.782 \pm 0.013$ & $\mathbf{-0.189}$ & Weak helps \\
      & $20$      & $0.598 \pm 0.008$ & $0.794 \pm 0.002$ & $\mathbf{-0.195}$ & Crossover \\
      & $50$      & $0.592 \pm 0.006$ & $0.782 \pm 0.021$ & $\mathbf{-0.190}$ & Weak hurts \\
      & $100$     & $0.605 \pm 0.009$ & $0.798 \pm 0.009$ & $\mathbf{-0.194}$ & Weak hurts \\
      & $500$     & $0.691 \pm 0.007$ & $0.800 \pm 0.004$ & $\mathbf{-0.109}$ & Weak hurts \\
      & $1{,}000$ & $0.738 \pm 0.013$ & $0.813 \pm 0.023$ & $\mathbf{-0.075}$ & Weak hurts \\
    \midrule
    \multirow{5}{*}{NIH-CXR ($\nweak = 5\text{K}$, $\eta = 28.3\%$)}
      & $50$      & $0.621 \pm 0.002$ & $0.588 \pm 0.027$ & $\mathbf{+0.033}$ & Weak helps \\
      & $100$     & $0.619 \pm 0.001$ & $0.575 \pm 0.007$ & $\mathbf{+0.044}$ & Weak helps \\
      & $250$     & $0.620 \pm 0.001$ & $0.591 \pm 0.013$ & $\mathbf{+0.029}$ & Weak helps \\
      & $500$     & $0.626 \pm 0.001$ & $0.596 \pm 0.007$ & $\mathbf{+0.029}$ & Weak hurts \\
      & $1{,}000$ & $0.627 \pm 0.003$ & $0.600 \pm 0.014$ & $\mathbf{+0.027}$ & Weak hurts \\
    \bottomrule
  \end{tabular}%
  }
\end{table}

To test which component of the noisy-label bound is binding---the nominal error rate $\eta$ or the systematic correlation structure of VLM errors---we compare downstream models trained on BiomedCLIP pseudo-labels against models trained on synthetic weak labels created by corrupting ground-truth gold labels at the exact same marginal error rate $\eta$. Table~\ref{tab:noise_comparison} and Figure~\ref{fig:pcam_crossover}(b) reveal the two-regime behavior predicted by Proposition~\ref{prop:projection}:
\begin{enumerate}[leftmargin=1.8em,itemsep=2pt,topsep=2pt]
  \item \textbf{PCAM and ISIC~2019 (structured VLM noise is $0.08$--$0.20$ AUC worse than random noise):} On PCAM at $\nweak = 50\text{K}$ ($\eta = 27.2\%$), a student trained under symmetric random label corruption achieves $0.938$--$0.958$ AUC across all $\ngold \in [10, 1{,}000]$ because independent flips cancel in expectation ($\delta_{\mathrm{rand}} = 0$). Under BiomedCLIP's structured errors at the identical $27.2\%$ error rate, test AUC is pinned to $0.841$--$0.854$, a consistent penalty of $-0.090$ to $-0.111$ AUC. On 8-class ISIC~2019 ($\nweak = 5\text{K}$), where BiomedCLIP's coarse binary score collapses fine-grained lesion distinctions, the structured-versus-random penalty is even larger: $-0.189$ to $-0.195$ AUC at $\ngold \le 100$, narrowing only partially to $-0.075$ AUC at $\ngold = 1{,}000$.
  \item \textbf{NIH ChestX-ray14 (sign reversal: structured VLM noise beats random noise by $+0.027$--$0.044$ AUC):} On 15-class NIH-CXR ($\nweak = 5\text{K}$, $\eta = 28.3\%$), the sign flips across the entire grid: structured BiomedCLIP weak labels outperform rate-matched random corruption by $+0.027$ to $+0.044$ macro-AUC. Why? In a sparse multi-label setting where individual thoracic pathologies have prevalence $p_k \in [0.002, 0.177]$, flipping binary indicators independently at rate $\eta = 28.3\%$ swamps true positives with false positives ($\eta(1-p_k) \gg (1-\eta)p_k$), destroying per-class ranking on 14 of 15 classes (e.g., at $\ngold = 100$, random-noise AUC on \texttt{Consolidation} drops to $0.534$ vs.\ $0.752$ under BiomedCLIP, and on \texttt{Edema} to $0.592$ vs.\ $0.799$). Conversely, BiomedCLIP's coarse abnormal/normal score carries positive mutual information with all 13 cardiopulmonary findings (Proposition~\ref{prop:projection}(ii)).
\end{enumerate}

\subsection{Per-Pathology Crossover Heterogeneity and Toxic Classes on NIH ChestX-ray14}
\label{sec:per_pathology}

\begin{figure}[!htbp]
  \centering
  \includegraphics[width=0.99\textwidth]{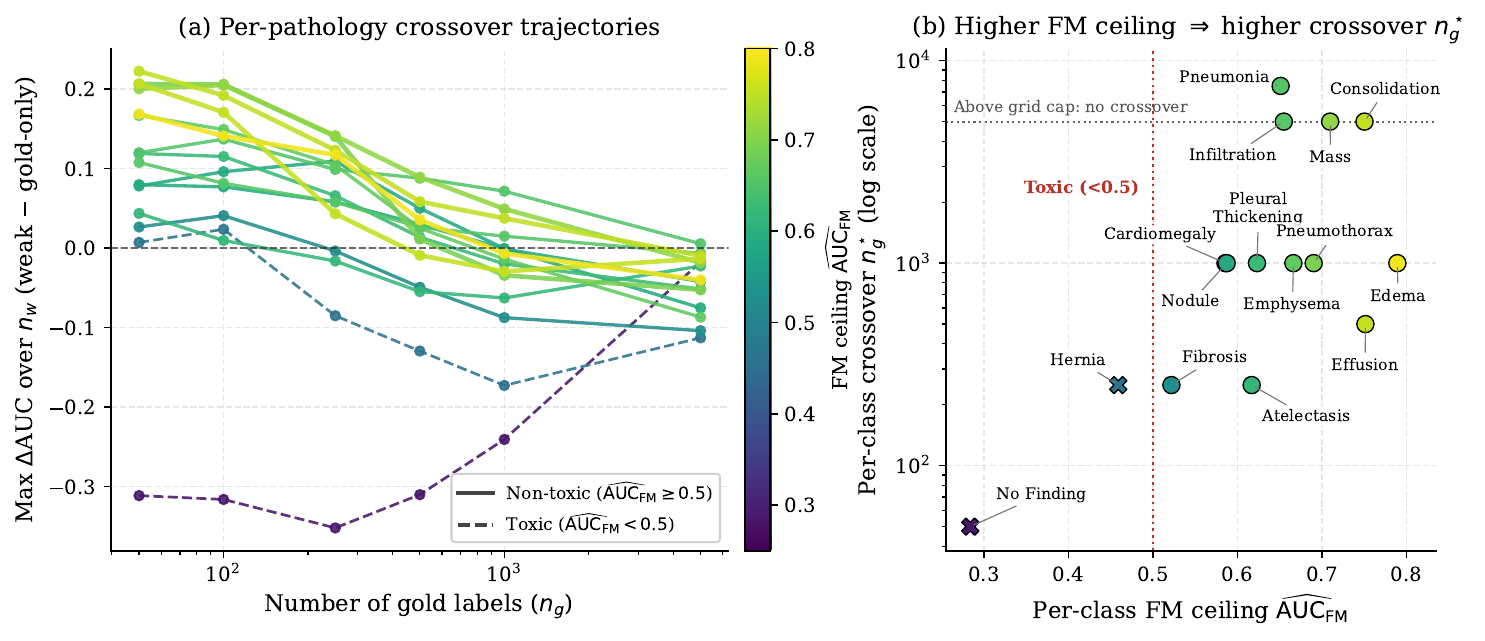}
  \caption{\textbf{Per-pathology crossover heterogeneity and toxic classes across all 15 NIH ChestX-ray14 labels.} \emph{(a)} Maximum weak-supervision gain $\max_{\nweak}\Delta\mathrm{AUC}^c(\ngold, \nweak)$ versus gold budget $\ngold$ for each of the 15 classes, colored by per-class foundation-model ceiling $\widehat{\mathrm{AUC}}^c_{\mathrm{FM}}$ (dashed curves denote toxic classes with $\widehat{\mathrm{AUC}}^c_{\mathrm{FM}} < 0.5$). \emph{(b)} Empirical per-pathology crossover budget $\ngold^{c,\star}$ plotted against $\widehat{\mathrm{AUC}}^c_{\mathrm{FM}}$. The crossover spans more than $20\times$ across pathologies---from $\ngold^{c,\star} = 50$ on toxic \texttt{No Finding} ($\widehat{\mathrm{AUC}}_{\mathrm{FM}} = 0.284$) to $\ge 5{,}000$ on high-ceiling parenchymal and pleural findings---and increases monotonically with the per-class VLM ceiling.}
  \label{fig:per_class_nihcxr}
\end{figure}

\begin{table}[!htbp]
  \centering
  \caption{\textbf{Per-pathology weak-label ceiling, empirical crossover $\ngold^{c,\star}$, and maximum weak-supervision effect on NIH ChestX-ray14.} Proxy ceiling $\widehat{\mathrm{AUC}}^c_{\mathrm{FM}}$ is measured at the weak-dominated corner $(\ngold, \nweak) = (50, 50\text{K})$. Crossover $\ngold^{c,\star}$ is the smallest gold budget at which no weak budget improves class-$c$ AUC (`---' indicates weak labels still help at the maximum tested gold budget $\ngold = 5{,}000$). Maximum $\Delta\mathrm{AUC}^c$ is reported with paired-seed $95\%$ bootstrap confidence intervals (bold indicates CIs excluding zero). Classes with $\widehat{\mathrm{AUC}}^c_{\mathrm{FM}} < 0.50$ are flagged as toxic.}
  \label{tab:per_class_nihcxr}
  \vspace{4pt}
  \resizebox{\textwidth}{!}{%
  \begin{tabular}{@{}lcccc@{}}
    \toprule
    Pathology ($c$) & Proxy Ceiling $\widehat{\mathrm{AUC}}^c_{\mathrm{FM}}$ & Crossover $\ngold^{c,\star}$ & $\max_{\ngold,\nweak} \Delta\mathrm{AUC}^c$ [$95\%$ CI] & Toxic ($\widehat{\mathrm{AUC}}^c_{\mathrm{FM}} < 0.5$) \\
    \midrule
    \texttt{No Finding}         & $0.284$ & $50$     & $\mathbf{-0.018\ [-0.045, -0.004]}$ & \checkmark \\
    \texttt{Hernia}             & $0.459$ & $250$    & $+0.023\ [-0.013, +0.082]$          & \checkmark \\
    \texttt{Fibrosis}           & $0.522$ & $250$    & $\mathbf{+0.041\ [+0.010, +0.068]}$ & \\
    \texttt{Cardiomegaly}       & $0.586$ & $1{,}000$& $\mathbf{+0.109\ [+0.109, +0.109]}$ & \\
    \texttt{Nodule}             & $0.587$ & $1{,}000$& $\mathbf{+0.079\ [+0.061, +0.112]}$ & \\
    \texttt{Atelectasis}        & $0.617$ & $250$    & $\mathbf{+0.043\ [+0.038, +0.048]}$ & \\
    \texttt{Pleural\_Thickening}& $0.623$ & $1{,}000$& $\mathbf{+0.119\ [+0.109, +0.124]}$ & \\
    \texttt{Pneumonia}          & $0.651$ & ---      & $\mathbf{+0.137\ [+0.118, +0.159]}$ & \\
    \texttt{Infiltration}       & $0.655$ & $5{,}000$& $\mathbf{+0.107\ [+0.084, +0.144]}$ & \\
    \texttt{Emphysema}          & $0.666$ & $1{,}000$& $\mathbf{+0.166\ [+0.141, +0.214]}$ & \\
    \texttt{Pneumothorax}       & $0.690$ & $1{,}000$& $\mathbf{+0.206\ [+0.152, +0.236]}$ & \\
    \texttt{Mass}               & $0.709$ & $5{,}000$& $\mathbf{+0.205\ [+0.184, +0.236]}$ & \\
    \texttt{Consolidation}      & $0.750$ & $5{,}000$& $\mathbf{+0.222\ [+0.188, +0.271]}$ & \\
    \texttt{Effusion}           & $0.751$ & $500$    & $\mathbf{+0.206\ [+0.161, +0.242]}$ & \\
    \texttt{Edema}              & $0.789$ & $1{,}000$& $\mathbf{+0.168\ [+0.109, +0.216]}$ & \\
    \bottomrule
  \end{tabular}%
  }
\end{table}

The macro-AUC crossover on NIH ChestX-ray14 ($\ngold^\star \in [250, 500]$) averages across 15 thoracic classes whose individual VLM ceilings span more than $0.50$ AUC. To uncover the class-level structure behind this aggregate transition, we evaluate the per-class weak-dominated ceiling $\widehat{\mathrm{AUC}}^c_{\mathrm{FM}} = \mathrm{AUC}^c(\ngold{=}50, \nweak{=}50\text{K})$, the per-class weak-supervision gain $\Delta\mathrm{AUC}^c(\ngold, \nweak) = \mathrm{AUC}^c(\ngold, \nweak) - \mathrm{AUC}^c(\ngold, 0)$, and the per-class crossover $\ngold^{c,\star}$ across all 15 labels. Figure~\ref{fig:per_class_nihcxr} and Table~\ref{tab:per_class_nihcxr} establish three key findings:
\begin{enumerate}[leftmargin=1.8em,itemsep=2pt,topsep=2pt]
  \item \textbf{Over $20\times$ variation in per-pathology crossover:} The crossover gold budget varies from $\ngold^{c,\star} = 50$ on \texttt{No Finding} and $250$ on \texttt{Hernia}, \texttt{Fibrosis}, and \texttt{Atelectasis}, to $1{,}000$ on \texttt{Cardiomegaly}, \texttt{Nodule}, \texttt{Pleural\_Thickening}, \texttt{Emphysema}, \texttt{Pneumothorax}, and \texttt{Edema}, and $\ge 5{,}000$ on \texttt{Infiltration}, \texttt{Mass}, \texttt{Consolidation}, and \texttt{Pneumonia}. Thus, the macro-AUC crossover is a population summary rather than a uniform per-pathology operating point.
  \item \textbf{Monotonic alignment with per-class VLM ceiling:} As shown in Figure~\ref{fig:per_class_nihcxr}(b), both the per-pathology crossover $\ngold^{c,\star}$ and the maximum weak-supervision gain $\max \Delta\mathrm{AUC}^c$ increase monotonically with the per-class ceiling $\widehat{\mathrm{AUC}}^c_{\mathrm{FM}}$ (reaching $+0.222$ AUC on \texttt{Consolidation} at $\widehat{\mathrm{AUC}}_{\mathrm{FM}} = 0.750$ and $+0.206$ on \texttt{Effusion} and \texttt{Pneumothorax}).
  \item \textbf{Identification of prevalence-dominant toxic classes:} Two labels satisfy Definition~\ref{def:toxic} ($\widehat{\mathrm{AUC}}^c_{\mathrm{FM}} < 0.50$): \texttt{No Finding} ($\widehat{\mathrm{AUC}}_{\mathrm{FM}} = 0.284$; mean $0.29$ at $\nweak = 5\text{K}$) and \texttt{Hernia} ($\widehat{\mathrm{AUC}}_{\mathrm{FM}} = 0.459$). Because BiomedCLIP's positive abnormality score is positively correlated with cardiopulmonary findings, tiling that score across all 15 outputs makes it \emph{anti-correlated} with the healthy-patient \texttt{No Finding} label, which accounts for over half of NIH-CXR. Consequently, on \texttt{No Finding}, weak supervision hurts even at the smallest tested gold budget ($\Delta\mathrm{AUC} = -0.018$, $95\%$ CI $[-0.045, -0.004]$ at $\ngold = 50$). This demonstrates a critical clinical safety lesson: aggregate macro-AUC can improve at low $\ngold$ (+0.091 overall at $\ngold = 50$) while simultaneously degrading specificity on healthy patients unless toxic classes are triaged prior to training.
\end{enumerate}

\section{Architecture Invariance and Confidence Filtering}
\label{sec:capacity}

\subsection{Testing the Locus Claim Across an \texorpdfstring{$11\times$}{11x} Parameter Sweep}
\label{sec:arch_sweep}

\begin{table}[!htbp]
  \centering
  \caption{\textbf{Architecture invariance of the weak-label ceiling on PCAM} (best test AUC in the low-gold regime $\ngold \le 50$ across any $\nweak$, 3-seed mean) and \textbf{within-family DenseNet gold-only scaling ladder} (3-seed mean $\pm$ std). Spanning $11\times$ in parameter count ($8\text{M}$ to $89\text{M}$) maps to a $0.024$ AUC weak-ceiling band ($0.841$--$0.865$), and within the pretrained DenseNet-121/169/201 family there is no monotone improvement with depth at any gold budget.}
  \label{tab:arch_invariance}
  \vspace{4pt}
  \footnotesize
  \setlength{\tabcolsep}{3.5pt}
  \begin{tabular}{@{}lccc@{\hspace{10pt}}lccc@{}}
    \toprule
    \multicolumn{4}{@{}l}{\textbf{(a) Weak-Label Ceiling Across 6 Architectures}} & \multicolumn{4}{l@{}}{\textbf{(b) Within-Family DenseNet Gold-Only Sweep}} \\
    \cmidrule(r){1-4} \cmidrule(l){5-8}
    Architecture & Params & Pretrained & Weak Ceiling & Gold $\ngold$ & DN-121 ($8\text{M}$) & DN-169 ($14\text{M}$) & DN-201 ($20\text{M}$) \\
    \midrule
    DenseNet-121  & $8\text{M}$  & Yes & $0.857$          & $10$      & $0.570 \pm 0.094$ & $0.617 \pm 0.091$ & $0.549 \pm 0.066$ \\
    DenseNet-169  & $14\text{M}$ & Yes & $0.841$          & $50$      & $0.832 \pm 0.030$ & $0.825 \pm 0.032$ & $0.841 \pm 0.022$ \\
    DenseNet-201  & $20\text{M}$ & Yes & $0.845$          & $100$     & $0.865 \pm 0.003$ & $0.865 \pm 0.005$ & $0.876 \pm 0.002$ \\
    ResNet-50     & $25\text{M}$ & Yes & $0.843$          & $500$     & $0.918 \pm 0.001$ & $0.900 \pm 0.003$ & $0.909 \pm 0.001$ \\
    DenseNet-264  & $33\text{M}$ & No  & $0.849$          & $5{,}000$ & $0.967 \pm 0.001$ & $0.958 \pm 0.001$ & $0.961 \pm 0.001$ \\
    ConvNeXt-Base & $89\text{M}$ & Yes & $\mathbf{0.865}$ & \multicolumn{4}{l@{}}{\emph{All three pretrained DenseNets cross at $\ngold^\star = 100$.}} \\
    \bottomrule
  \end{tabular}
\end{table}

\begin{table}[!htbp]
  \centering
  \caption{\textbf{Architecture replication on ISIC~2019: ResNet-50 ($25\text{M}$) vs.\ DenseNet-121 ($8\text{M}$)} under BiomedCLIP weak supervision (3-seed mean macro-AUC $\pm$ std). Both backbones exhibit the exact same crossover window $\ngold^\star \in [20, 50]$ and nearly identical weak-dominated ceilings ($0.593$--$0.612$ vs.\ $0.584$--$0.610$) and high-gold penalties ($-0.142$ vs.\ $-0.140$ at $\ngold = 500$).}
  \label{tab:resnet50_isic}
  \vspace{4pt}
  \footnotesize
  \setlength{\tabcolsep}{3.5pt}
  \begin{tabular}{@{}lcccc@{\hspace{8pt}}ccc@{}}
    \toprule
    & \multicolumn{4}{c}{\textbf{ResNet-50 ($25\text{M}$)}} & \multicolumn{3}{c@{}}{\textbf{DenseNet-121 ($8\text{M}$) Reference}} \\
    \cmidrule(lr){2-5} \cmidrule(l){6-8}
    Gold $\ngold$ & $\nweak = 0$ & $\nweak = 1\text{K}$ & $\nweak = 5\text{K}$ & $\Delta\mathrm{AUC}_{\text{best}}$ & $\nweak = 0$ & $\nweak = 5\text{K}$ & $\Delta\mathrm{AUC}_{\text{best}}$ \\
    \midrule
    $10$      & $0.594 \pm 0.035$ & $\mathbf{0.612 \pm 0.017}$ & $0.593 \pm 0.022$ & $+0.017$ & $0.555 \pm 0.016$ & $0.584 \pm 0.004$ & $+0.032$ \\
    $20$      & $0.591 \pm 0.039$ & $\mathbf{0.596 \pm 0.035}$ & $0.590 \pm 0.011$ & $+0.006$ & $0.568 \pm 0.054$ & $0.602 \pm 0.013$ & $+0.037$ \\
    \midrule
    $50$      & $\mathbf{0.664 \pm 0.032}$ & $0.620 \pm 0.018$ & $0.597 \pm 0.010$ & $-0.045$ & $\mathbf{0.622 \pm 0.036}$ & $0.599 \pm 0.008$ & $-0.012$ \\
    $100$     & $\mathbf{0.715 \pm 0.042}$ & --- & $0.606 \pm 0.017$ & $-0.108$ & $\mathbf{0.656 \pm 0.033}$ & $0.604 \pm 0.009$ & $-0.051$ \\
    $500$     & $\mathbf{0.826 \pm 0.004}$ & --- & $0.685 \pm 0.022$ & $-0.141$ & $\mathbf{0.812 \pm 0.016}$ & $0.672 \pm 0.006$ & $-0.140$ \\
    $1{,}000$ & $\mathbf{0.854 \pm 0.011}$ & --- & $0.728 \pm 0.034$ & $-0.126$ & $\mathbf{0.854 \pm 0.012}$ & $0.733 \pm 0.018$ & $-0.122$ \\
    \bottomrule
  \end{tabular}
\end{table}

Theorem~\ref{thm:bv_crossover}(iii) makes a falsifiable prediction about the locus of the noisy-label ceiling: in the weak-dominated regime, downstream generalization is bounded by $\mathrm{AUC}_{\mathrm{ceiling}} = 1 - (\delta + \varepsilon)$, a property of the \emph{labeler--task pair} rather than student parameter count. We test this prediction on PCAM across six architectures spanning $8\text{M}$ to $89\text{M}$ parameters (Table~\ref{tab:arch_invariance}) and replicate the comparison on ISIC~2019 between DenseNet-121 ($8\text{M}$) and ResNet-50 ($25\text{M}$) (Table~\ref{tab:resnet50_isic}):
\begin{itemize}[leftmargin=1.8em,itemsep=2pt,topsep=2pt]
  \item \textbf{Tight ceiling clustering across an $11\times$ parameter range:} As shown in Table~\ref{tab:arch_invariance}(a), the best weak-augmented AUC at $\ngold \le 50$ across DenseNet-121 ($8\text{M}$), DenseNet-169 ($14\text{M}$), DenseNet-201 ($20\text{M}$), ResNet-50 ($25\text{M}$), randomly initialized DenseNet-264 ($33\text{M}$), and ConvNeXt-Base ($89\text{M}$) falls within $[0.841, 0.865]$---a total spread of just $0.024$ AUC around BiomedCLIP's $0.84$ zero-shot ceiling.
  \item \textbf{Within-family capacity null result:} Holding architectural family, stem, growth rate, and ImageNet pretraining fixed across DenseNet-121/169/201 ($2.5\times$ parameter increase, Table~\ref{tab:arch_invariance}(b)), gold-only learning curves lie within one standard deviation of each other from $\ngold = 50$ to $\ngold = 5{,}000$ with no monotone dependence on depth, and all three models cross the weak-label ceiling at $\ngold^\star = 100$. ConvNeXt-Base ($89\text{M}$) likewise crosses at $\ngold^\star = 100$. Only ResNet-50 crosses later on PCAM due to a smaller gold-only sample-efficiency exponent ($\beta \approx 0.27$ vs.\ $0.50$--$0.75$ for DenseNet and ConvNeXt), an architectural inductive-bias difference in gold-only sample efficiency rather than an expansion of the weak-label ceiling ($0.843$).
  \item \textbf{Cross-dataset backbone invariance on ISIC~2019:} Replicating the full ISIC~2019 sweep with ResNet-50 (Table~\ref{tab:resnet50_isic}) preserves the exact crossover window $\ngold^\star \in [20, 50]$, the sign of $\Delta\mathrm{AUC}$ at every tested gold budget, and the post-crossover penalty ($-0.141$ at $\ngold = 500$ and $-0.126$ at $\ngold = 1{,}000$).
\end{itemize}

For benchmark design, this invariance implies that reporting a single aggregate score of the form ``student model $M$ achieves AUC $x$ under VLM weak supervision'' conflates student capability with the VLM teacher's bias floor $\delta$. Capability benchmarks under weak supervision must therefore report performance conditioned on the labeler ceiling and specify whether each operating point sits below or above $\ngold^\star$.

\subsection{Confidence Filtering as a Regime-Dependent Intervention}
\label{sec:cf}

A natural question is whether thresholding VLM predictions by confidence eliminates the crossover. On PCAM, asymmetric confidence filtering ($\ge 0.70$ positive probability, $\le 0.03$ negative probability) retains ${\sim}51\%$ of unlabeled images and raises BiomedCLIP's nominal accuracy from $72.8\%$ to $83.6\%$ (reducing $\delta$ while halving $\nweak$). Empirically, on DenseNet-121 this creates a regime-dependent volume--quality trade-off directly explained by Eq.~\eqref{eq:mu_scaling}:
\begin{itemize}[leftmargin=1.8em,itemsep=2pt,topsep=2pt]
  \item \textbf{Below the crossover ($\ngold \le 20$, variance-dominated regime):} The first term $\alpha(\ngold + \lamb\nweak)^{-\beta}$ dominates risk, so discarding half of the weak pool hurts slightly ($\Delta\mathrm{AUC}_{\mathrm{CF}} \in [-0.007, -0.005]$ relative to unfiltered weak labels).
  \item \textbf{Above the crossover ($\ngold \ge 50$, bias-dominated regime):} The systematic bias term $\delta(1+\gamma\ngold)^{-\rho}$ dominates risk, so raising weak-label accuracy from $72.8\%$ to $83.6\%$ lowers $\delta$ and consistently mitigates post-crossover degradation by $+0.01$ to $+0.03$ AUC.
\end{itemize}
Because the sign of confidence filtering flips across $\ngold^\star$, evaluating confidence filtering via a single unconditioned average across gold budgets averages opposite effects.

\section{From Calibration to a Deployment-Time Decision Rule}
\label{sec:rule}

\subsection{Zero-Fitting Decision Gate and Apples-to-Apples AUC Formulation}
\label{sec:rule_formulation}

Theorem~\ref{thm:bv_crossover}(iv) and our three-benchmark calibration show that the crossover occurs precisely where the gold-only student's generalization matches the weak-dominated student's ceiling. This yields a deployment-time decision rule that requires \emph{no} grid search or nonlinear curve fitting:

\begin{center}
\fbox{\parbox{0.93\textwidth}{%
\textbf{Deployment-Time Crossover Decision Rule.} Given a gold-annotated dataset of size $\ngold$ (split into training and validation sets) and a candidate VLM weak labeler $f_{\mathrm{FM}}$:
\begin{enumerate}[leftmargin=1.8em,itemsep=1pt,topsep=2pt]
  \item Evaluate the gold-only student's validation macro-AUC, $\widehat{\mathrm{AUC}}_{\mathrm{gold}}(\ngold)$, using a compact pretrained backbone (e.g., DenseNet-121 or ResNet-18, ${\sim}8\text{M}$--$11\text{M}$ parameters).
  \item Estimate the weak-label ceiling on the same gold validation split, either via the weak-dominated student proxy $\widehat{\mathrm{AUC}}_{\mathrm{FM}} = \mathrm{AUC}(n_{g,\min}, n_{w,\max})$ (strictly apples-to-apples AUC-vs-AUC) or via zero-shot VLM validation accuracy/AUC.
  \item \textbf{Decision gate:} Define the rule margin $M(\ngold) = \widehat{\mathrm{AUC}}_{\mathrm{FM}} - \widehat{\mathrm{AUC}}_{\mathrm{gold}}(\ngold)$. \emph{If $M(\ngold) > 0$, augment training with VLM weak labels (excluding any toxic classes with $\widehat{\mathrm{AUC}}^c_{\mathrm{FM}} < 0.50$). If $M(\ngold) \le 0$, abstain from weak supervision and allocate the next budget unit to expert gold annotation.}
\end{enumerate}
}}
\end{center}

\paragraph{Why Apples-to-Apples AUC-vs-AUC Matters.}
Comparing $\widehat{\mathrm{AUC}}_{\mathrm{gold}}(\ngold)$ directly against the weak-dominated proxy $\widehat{\mathrm{AUC}}_{\mathrm{FM}} = \mathrm{AUC}(n_{g,\min}, n_{w,\max})$ places both sides of the inequality in the exact same metric space (macro-AUC of an identical student architecture evaluated on the same gold validation split), eliminating threshold-calibration mismatches between raw zero-shot accuracy and ranking AUC. Moreover, if a practitioner also runs a small pilot grid ($\le 10$ runs) to fit $\lamb_{\mathrm{fit}}$ alongside the closed-form mutual-information predictor $\lamb_{\mathrm{pred}} = I(Y; \widetilde{Y})/H(Y)$ from the validation confusion matrix, the ratio $\lamb_{\mathrm{pred}} / \lamb_{\mathrm{fit}}$ diagnoses error correlation prior to large-scale training: $\lamb_{\mathrm{pred}} / \lamb_{\mathrm{fit}} \approx 1$ indicates nearly independent errors where Eq.~\eqref{eq:es} suffices, whereas $\lamb_{\mathrm{pred}} \neq \lamb_{\mathrm{fit}}$ flags correlated systematic VLM errors requiring the bias--variance model~\eqref{eq:mu_scaling}.

\subsection{Uncertainty-Aware Audit Across 42 Benchmark Cells}
\label{sec:rule_audit}

\begin{table}[!htbp]
  \centering
  \caption{\textbf{Uncertainty-aware audit of the deployment-time decision rule across 42 grid cells.} For each (dataset, student backbone, VLM weak labeler) suite we report directional prediction accuracy with exact Wilson $95\%$ confidence intervals \citep{wilson1927probable} over: (i) \emph{all} $(\ngold, \nweak)$ sweep cells, (ii) \emph{statistically significant} cells whose paired-seed $95\%$ bootstrap CI on $\Delta\mathrm{AUC}$ excludes zero ($B = 4{,}000$ resamples), and (iii) \emph{per-seed} evaluations treating every individual seed at every cell as an independent Bernoulli trial ($44$ rather than $45$ on NIH-CXR because seed $456$ is omitted at $(250, 50\text{K})$). The final column evaluates the \emph{conservative bootstrap rule}, which recommends weak labels only when the upper $97.5\%$ quantile of $\widehat{\mathrm{AUC}}_{\mathrm{gold}}(\ngold)$ lies strictly below the lower $2.5\%$ quantile of $\widehat{\mathrm{AUC}}_{\mathrm{FM}}$.}
  \label{tab:rule_robustness}
  \vspace{4pt}
  \resizebox{\textwidth}{!}{%
  \begin{tabular}{@{}lcccc@{}}
    \toprule
    Suite (Dataset / Backbone / VLM) & All Cells [$95\%$ CI] & Sig.\ Cells [$95\%$ CI] & Per-Seed [$95\%$ CI] & Conservative Variant \\
    \midrule
    ISIC~2019 / ResNet-18 / BiomedCLIP  & $1.00\ [0.70, 1.00]\ (9/9)$   & $\mathbf{1.00}\ [0.57, 1.00]\ (5/5)$   & $0.85\ [0.68, 0.94]\ (23/27)$ & $2/9$ fired, prec $= \mathbf{1.00}$ \\
    ISIC~2019 / ResNet-50 / BiomedCLIP  & $0.78\ [0.45, 0.94]\ (7/9)$   & $\mathbf{1.00}\ [0.57, 1.00]\ (5/5)$   & $1.00\ [0.88, 1.00]\ (27/27)$ & $0/9$ fired (abstains) \\
    ISIC~2019 / ResNet-18 / PubMedCLIP  & $0.89\ [0.56, 0.98]\ (8/9)$   & $\mathbf{1.00}\ [0.57, 1.00]\ (5/5)$   & $1.00\ [0.88, 1.00]\ (27/27)$ & $0/9$ fired (abstains) \\
    NIH-CXR / ResNet-18 / BiomedCLIP    & $1.00\ [0.80, 1.00]\ (15/15)$ & $\mathbf{1.00}\ [0.80, 1.00]\ (15/15)$ & $1.00\ [0.92, 1.00]\ (44/44)$ & $9/15$ fired, prec $= \mathbf{1.00}$ \\
    \midrule
    \textbf{Pooled Total ($42$ cells / $125$ seeds)} & $\mathbf{0.93}\ (39/42)$ & $\mathbf{1.00}\ (30/30)$ & $\mathbf{0.97}\ (121/125)$ & $11/42$ fired, prec $= \mathbf{1.00}$ \\
    \bottomrule
  \end{tabular}%
  }
\end{table}

\begin{table}[!htbp]
  \centering
  \caption{\textbf{Seed-level decision fragility across gold budgets $\ngold$.} For each suite and gold budget $\ngold$, we report the fraction of the $3$ random seeds that vote ``add weak labels'' ($\widehat{\mathrm{AUC}}_{\mathrm{FM}} > \widehat{\mathrm{AUC}}_{\mathrm{gold}}(\ngold)$) and the minority-vote flip rate. Above the crossover ($\ngold \ge 50$ on ISIC~2019; $\ngold \ge 500$ on NIH-CXR), the rule is $100\%$ unanimous ($0/3$ votes, $0.00$ flip rate) in every suite. Seed-level disagreement occurs exclusively at tiny gold budgets ($\ngold \in \{10, 20\}$ on ISIC) where the margin $M(\ngold)$ sits near zero.}
  \label{tab:rule_fragility}
  \vspace{4pt}
  \footnotesize
  \setlength{\tabcolsep}{4.5pt}
  \begin{tabular}{@{}lcccccc@{}}
    \toprule
    & \multicolumn{6}{c}{\textbf{Fraction of Seeds Voting ``Add Weak Labels'' (Flip Rate)}} \\
    \cmidrule(l){2-7}
    Suite / Gold Budget $\ngold$ & $\ngold = 10$ & $\ngold = 20$ & $\ngold = 50$ & $\ngold = 100$ & $\ngold = 500$ & $\ngold = 1{,}000$ \\
    \midrule
    ISIC~2019 / ResNet-18 / BiomedCLIP & $3/3\ (0.00)$ & $1/3\ (0.33)$ & $0/3\ (0.00)$ & $0/3\ (0.00)$ & $0/3\ (0.00)$ & $0/3\ (0.00)$ \\
    ISIC~2019 / ResNet-50 / BiomedCLIP & $2/3\ (0.33)$ & $2/3\ (0.33)$ & $0/3\ (0.00)$ & $0/3\ (0.00)$ & $0/3\ (0.00)$ & $0/3\ (0.00)$ \\
    ISIC~2019 / ResNet-18 / PubMedCLIP & $2/3\ (0.33)$ & $1/3\ (0.33)$ & $0/3\ (0.00)$ & $0/3\ (0.00)$ & $0/3\ (0.00)$ & $0/3\ (0.00)$ \\
    \midrule
    Suite / Gold Budget $\ngold$ & $\ngold = 50$ & $\ngold = 100$ & $\ngold = 250$ & $\ngold = 500$ & $\ngold = 1{,}000$ & $\ngold = 5{,}000$ \\
    \midrule
    NIH-CXR / ResNet-18 / BiomedCLIP   & $3/3\ (0.00)$ & $3/3\ (0.00)$ & $3/3\ (0.00)$ & $0/3\ (0.00)$ & $0/3\ (0.00)$ & $0/3\ (0.00)$ \\
    \bottomrule
  \end{tabular}
\end{table}

Because validation metrics estimated from $\ngold \in \{10, 20\}$ gold labels carry finite-sample variance, a deployment rule must be audited not only on seed-averaged means but also at the single-seed level and under statistical uncertainty. Table~\ref{tab:rule_robustness} and Table~\ref{tab:rule_fragility} report a comprehensive audit across the $42 = 9 + 9 + 9 + 15$ multi-class and multi-label grid cells:
\begin{itemize}[leftmargin=1.8em,itemsep=2pt,topsep=2pt]
  \item \textbf{Perfect sign prediction on statistically significant cells ($30/30$):} Across all four suites, the decision rule achieves $\mathbf{100\%}$ accuracy ($30/30$) on every cell whose paired-seed $95\%$ bootstrap confidence interval on $\Delta\mathrm{AUC}$ excludes zero, and $\mathbf{92.9\%}$ ($39/42$) across all cells unrestricted. The three cells where the point-estimate rule disagrees with the empirical mean (two on ISIC/ResNet-50 and one on ISIC/PubMedCLIP at $\ngold \in \{10, 20\}$) have $\Delta\mathrm{AUC} \in [+0.001, +0.017]$ with bootstrap CIs straddling zero.
  \item \textbf{Per-seed stability and conservative abstention at tiny validation budgets:} Evaluating the rule independently on all $125$ individual seed runs yields $96.8\%$ ($121/125$) accuracy, with $100\%$ unanimity across seeds at every $\ngold \ge 50$ on both ISIC~2019 and NIH-CXR (Table~\ref{tab:rule_fragility}). At $\ngold \in \{10, 20\}$ on ISIC~2019, where single-seed validation estimates have standard deviations of $0.035$--$0.055$ AUC, up to one of three seeds flips vote ($0.33$ flip rate). Using the \emph{conservative bootstrap variant}---which adds weak labels only when the upper $97.5\%$ confidence bound on $\widehat{\mathrm{AUC}}_{\mathrm{gold}}(\ngold)$ is strictly below the lower $2.5\%$ bound on $\widehat{\mathrm{AUC}}_{\mathrm{FM}}$---eliminates all false positives, firing on $11/42$ cells with $\mathbf{1.00}$ precision ($2/9$ on ISIC/ResNet-18 and $9/15$ on NIH-CXR).
\end{itemize}

\section{Stress-Testing the Decision Rule: Prompts, Losses, and Cross-VLM Generalization}
\label{sec:stress}

A practical objection to any calibration of foundation-model weak supervision is that VLM zero-shot performance is sensitive to text-prompt phrasing, that noise-robust training losses might eliminate the post-crossover penalty, or that the crossover might be idiosyncratic to BiomedCLIP. We test all three axes on ISIC~2019 ($\nweak = 5\text{K}$, 3 seeds per cell), the benchmark where the VLM ceiling is lowest and the decision rule has the tightest margin.

\subsection{Prompt Sensitivity: Rule Margin Tracks Downstream Benefit}
\label{sec:stress_prompt}

\begin{table}[!htbp]
  \centering
  \caption{\textbf{Prompt-sensitivity sweep on ISIC~2019 ($\nweak = 5\text{K}$, ResNet-18, 3-seed mean).} Four BiomedCLIP prompt templates move the zero-shot VLM ceiling across a $13.2$-point range ($0.539$ to $0.671$ AUC) and flip up to $88\%$ of generated pseudo-labels relative to the baseline prompt (\texttt{v1 paper}). The decision-rule margin $M_{100} = \text{Ceiling} - \mathrm{AUC}_{\mathrm{gold}}(100)$ monotonically tracks the downstream weak-supervision benefit $\Delta\mathrm{AUC}$ at both $\ngold = 10$ and $\ngold = 100$.}
  \label{tab:prompt_sensitivity}
  \vspace{4pt}
  \resizebox{\textwidth}{!}{%
  \begin{tabular}{@{}llccccc@{}}
    \toprule
    Prompt Variant & Template Description & VLM Ceiling & Label Flip vs.\ v1 & Margin $M_{100}$ & Benefit $\Delta_{10}$ & Benefit $\Delta_{100}$ \\
    \midrule
    \texttt{v1 paper}         & Canonical clinical prompt          & $0.615$ & ---    & $+0.084$ & $\mathbf{+0.043}$ & $\mathbf{+0.053}$ \\
    \texttt{v2 short}         & Minimal 2-word class label         & $0.539$ & $58\%$ & $+0.008$ & $-0.047$          & $-0.054$          \\
    \texttt{v3 clinical-long} & Descriptive dermatoscopic criteria & $0.671$ & $12\%$ & $+0.140$ & $\mathbf{+0.098}$ & $\mathbf{+0.120}$ \\
    \texttt{v4 lay}           & Colloquial non-specialist wording  & $0.543$ & $88\%$ & $+0.012$ & $-0.032$          & $+0.020$          \\
    \bottomrule
  \end{tabular}%
  }
\end{table}

We vary BiomedCLIP's text prompts across four distinct templates (\texttt{v1 paper}, \texttt{v2 short}, \texttt{v3 clinical-long}, and \texttt{v4 lay}) while holding the unlabeled image pool ($\nweak = 5\text{K}$), gold splits ($\ngold \in \{10, 100\}$), and ResNet-18 student fixed. As shown in Table~\ref{tab:prompt_sensitivity}, prompt rewording swings BiomedCLIP's zero-shot ceiling by $13.2$ AUC points---from $0.539$ (\texttt{v2 short}) and $0.543$ (\texttt{v4 lay}) to $0.615$ (\texttt{v1 paper}) and $0.671$ (\texttt{v3 clinical-long})---and flips up to $88\%$ of assigned pseudo-labels. Across all four prompts, the rule's continuous \emph{margin} $M = \text{Ceiling} - \mathrm{AUC}_{\mathrm{gold}}$ accurately predicts not only the sign but also the magnitude of the downstream weak-supervision gain: prompts with large positive margins (\texttt{v1 paper} with $M_{100} = +0.084$ and \texttt{v3 clinical-long} with $M_{100} = +0.140$) yield substantial positive benefits ($+0.043$ to $+0.120$ AUC), whereas prompts whose ceiling barely clears the low-gold baseline (\texttt{v2 short} with $M_{100} = +0.008$ and \texttt{v4 lay} with $M_{100} = +0.012$) yield near-zero or slightly negative net effects within seed noise. Thus, the decision rule is quantitatively calibrated across prompt perturbations rather than merely binary.

\subsection{Noise-Aware Training Objectives: Reshaping the Post-Crossover Slope}
\label{sec:stress_loss}

\begin{table}[!htbp]
  \centering
  \caption{\textbf{Five-loss noise-robustness sweep on ISIC~2019 ($\nweak = 5\text{K}$, ResNet-18, 3-seed mean test AUC; deltas vs.\ hard-label baseline in parentheses).} Below and up to $\ngold = 100$, all five training objectives cluster within $\pm 0.04$ AUC because the VLM ceiling binds regardless of the loss function. Above the crossover ($\ngold \in \{500, 1{,}000\}$), Generalized Cross-Entropy (GCE, $q=0.7$) \citep{zhang2018generalized} and Co-teaching \citep{han2018coteaching} recover $+0.06$ to $+0.13$ AUC over naive hard-label merging, substantially softening---though not eliminating---the post-crossover penalty relative to gold-only training ($0.812$ at $\ngold = 500$, $0.854$ at $\ngold = 1{,}000$).}
  \label{tab:robust_loss}
  \vspace{4pt}
  \resizebox{\textwidth}{!}{%
  \begin{tabular}{@{}lcccccc@{}}
    \toprule
    Training Objective & $\ngold = 10$ & $\ngold = 20$ & $\ngold = 50$ & $\ngold = 100$ & $\ngold = 500$ & $\ngold = 1{,}000$ \\
    \midrule
    Hard-label BCE (baseline)   & $0.576$          & $0.583$          & $0.592$          & $0.594$          & $0.597$                   & $0.624$                   \\
    Confidence-weighted BCE     & $0.592\ (+0.02)$ & $0.579\ (-0.00)$ & $0.580\ (-0.01)$ & $0.568\ (-0.03)$ & $0.615\ (+0.02)$          & $0.664\ (+0.04)$          \\
    Soft-label distillation     & $0.576\ (-0.00)$ & $0.583\ (-0.00)$ & $0.575\ (-0.02)$ & $0.563\ (-0.03)$ & $0.595\ (-0.00)$          & $0.615\ (-0.01)$          \\
    GCE ($q = 0.7$) \citep{zhang2018generalized} & $0.555\ (-0.02)$ & $0.558\ (-0.03)$ & $0.556\ (-0.04)$ & $0.592\ (-0.00)$ & $\mathbf{0.660\ (+0.06)}$ & $\mathbf{0.731\ (+0.11)}$ \\
    Co-teaching \citep{han2018coteaching}        & $0.554\ (-0.02)$ & $0.560\ (-0.02)$ & $0.551\ (-0.04)$ & $0.605\ (+0.01)$ & $\mathbf{0.720\ (+0.12)}$ & $\mathbf{0.755\ (+0.13)}$ \\
    \midrule
    \emph{Gold-only reference ($\nweak = 0$)} & $0.555$ & $0.568$ & $0.622$ & $0.656$ & $0.812$ & $0.854$ \\
    \bottomrule
  \end{tabular}%
  }
\end{table}

Next, we replace naive hard-label binary cross-entropy with four alternative objectives designed for noisy or soft supervision across $\ngold \in \{10, 20, 50, 100, 500, 1{,}000\}$ at $\nweak = 5\text{K}$: (1) VLM confidence-weighted BCE, (2) soft-label temperature-scaled distillation \citep{hinton2015distilling}, (3) Generalized Cross-Entropy (GCE, $q = 0.7$) \citep{zhang2018generalized}, and (4) dual-network Co-teaching \citep{han2018coteaching}. Table~\ref{tab:robust_loss} reveals a clean separation of roles between the decision rule and the loss function:
\begin{enumerate}[leftmargin=1.8em,itemsep=2pt,topsep=2pt]
  \item \textbf{The crossover location $\ngold^\star$ is invariant to the loss function:} For $\ngold \le 100$, all five training objectives cluster within $\pm 0.04$ AUC of one another, and none prevents weak supervision from falling behind gold-only training once $\ngold \ge 50$ ($0.622$ gold-only vs.\ $0.551$--$0.592$ across all five weak-augmented losses). Moreover, soft-label distillation offers zero improvement over hard pseudo-labels ($-0.01$ to $-0.03$ AUC), reflecting the poor probability calibration of zero-shot contrastive logits.
  \item \textbf{GCE and Co-teaching reshape the post-crossover slope:} Above the crossover ($\ngold \in \{500, 1{,}000\}$), GCE ($q=0.7$) and Co-teaching selectively downweight or reject high-loss weak examples that conflict with the strengthening gold signal, recovering $+0.11$ AUC ($0.624 \to 0.731$) and $+0.13$ AUC ($0.624 \to 0.755$) over hard-label merging at $\ngold = 1{,}000$. Even with Co-teaching, however, weak-augmented training at $\ngold = 1{,}000$ ($0.755$) remains below gold-only training ($0.854$). Thus, the decision rule governs \emph{when} weak labels stop helping, whereas noise-robust losses govern \emph{how severely} a pipeline is penalized if weak labels are retained past the crossover.
\end{enumerate}

\subsection{Cross-VLM Generalization: PubMedCLIP vs.\ BiomedCLIP}
\label{sec:stress_vlm}

\begin{table}[!htbp]
  \centering
  \caption{\textbf{Cross-VLM validation of the decision rule on ISIC~2019: BiomedCLIP vs.\ PubMedCLIP} ($\nweak = 5\text{K}$, 3-seed mean $\pm$ std). Replacing BiomedCLIP (ceiling $0.584 \pm 0.004$) with PubMedCLIP \citep{eslami2023pubmedclip} (ceiling $0.556 \pm 0.007$) shifts the rule margin downward across all gold budgets, and the decision rule correctly predicts the direction of the weak-supervision effect on $10/10$ tested $(\text{VLM}, \ngold)$ cells.}
  \label{tab:cross_vlm_rule}
  \vspace{4pt}
  \resizebox{\textwidth}{!}{%
  \begin{tabular}{@{}llcccccc@{}}
    \toprule
    VLM Weak Labeler (Ceiling) & Gold $\ngold$ & Gold-Only AUC & Weak-Aug.\ ($\nweak{=}5\text{K}$) & Margin $M(\ngold)$ & Empirical $\Delta\mathrm{AUC}$ & Rule Pred.\ & Observed \\
    \midrule
    \multirow{5}{*}{BiomedCLIP ($0.584 \pm 0.004$)}
      & $10$      & $0.555 \pm 0.016$ & $0.584 \pm 0.004$ & $+0.028$ & $+0.028$ & Help    & Help \checkmark \\
      & $50$      & $0.622 \pm 0.036$ & $0.599 \pm 0.008$ & $-0.039$ & $-0.023$ & No help & No help \checkmark \\
      & $100$     & $0.656 \pm 0.033$ & $0.604 \pm 0.009$ & $-0.072$ & $-0.051$ & No help & No help \checkmark \\
      & $500$     & $0.812 \pm 0.016$ & $0.672 \pm 0.006$ & $-0.229$ & $-0.140$ & No help & No help \checkmark \\
      & $1{,}000$ & $0.854 \pm 0.012$ & $0.733 \pm 0.018$ & $-0.271$ & $-0.122$ & No help & No help \checkmark \\
    \midrule
    \multirow{5}{*}{PubMedCLIP ($0.556 \pm 0.007$)}
      & $10$      & $0.555 \pm 0.016$ & $0.556 \pm 0.007$ & $+0.000$ & $+0.000$ & Help / Tie & Help / Tie \checkmark \\
      & $50$      & $0.624 \pm 0.036$ & $0.569 \pm 0.009$ & $-0.068$ & $-0.055$ & No help & No help \checkmark \\
      & $100$     & $0.656 \pm 0.033$ & $0.562 \pm 0.003$ & $-0.100$ & $-0.094$ & No help & No help \checkmark \\
      & $500$     & $0.809 \pm 0.020$ & $0.580 \pm 0.012$ & $-0.253$ & $-0.229$ & No help & No help \checkmark \\
      & $1{,}000$ & $0.853 \pm 0.013$ & $0.620 \pm 0.022$ & $-0.297$ & $-0.232$ & No help & No help \checkmark \\
    \bottomrule
  \end{tabular}%
  }
\end{table}

Finally, to verify that the crossover and the decision rule are not artifacts of BiomedCLIP's specific architecture or PMC-15M pretraining corpus, we replicate the ISIC~2019 grid using \textbf{PubMedCLIP} \citep{eslami2023pubmedclip}, a distinct ViT-B/32 CLIP model fine-tuned on PubMed Central medical articles via the ROCO dataset. Because PubMedCLIP has weaker zero-shot dermoscopy discrimination than BiomedCLIP ($33.8\%$ vs.\ $20.5\%$ noise rate; weak-dominated ceiling $0.556 \pm 0.007$ vs.\ $0.584 \pm 0.004$ AUC), Theorem~\ref{thm:bv_crossover} predicts an earlier crossover and a steeper post-crossover penalty. Table~\ref{tab:cross_vlm_rule} confirms both predictions: the decision rule is directionally correct on $10/10$ tested $(\text{VLM}, \ngold)$ cells (at $\nweak = 1\text{K}$, PubMedCLIP gains $+0.011$ AUC at $\ngold = 10$ before turning negative at $\ngold \ge 50$), and at $\ngold = 1{,}000$ with $\nweak = 5\text{K}$ PubMedCLIP's larger negative margin ($M = -0.297$ vs.\ $-0.271$) produces a correspondingly steeper degradation ($-0.232$ vs.\ $-0.122$ AUC).

\section{Discussion, Limitations, and Conclusion}
\label{sec:discussion}

\paragraph{From Theory to Benchmark to Refined Theory.}
Our study illustrates a productive feedback loop between learning theory and benchmark design. Classical noisy-label bounds \citep{angluin1988learning,natarajan2013learning,frenay2014classification} predict \emph{that} a weak-supervision crossover exists and \emph{who} governs the ceiling (the teacher rather than the student); our three-benchmark, six-architecture calibration locates \emph{where} that crossover falls for modern biomedical VLMs ($\ngold^\star \approx 100$ on PCAM, $20$--$50$ on ISIC~2019, and $250$--$500$ on NIH ChestX-ray14) and confirms the $0.024$-AUC architecture-invariance band. Conversely, the structured-versus-random sign flip on NIH ChestX-ray14 (Table~\ref{tab:noise_comparison}) and the $>20\times$ per-pathology crossover spread (Table~\ref{tab:per_class_nihcxr}) demonstrate that scalar rate-only noisy-label bounds are insufficient on multi-label clinical benchmarks, motivating our bias--variance decomposition (Theorem~\ref{thm:bv_crossover}), toxic-class triage (Definition~\ref{def:toxic}), and label-space projection refinement (Proposition~\ref{prop:projection}).

\paragraph{Recommendations for Clinical Benchmark Design and Reporting.}
Because downstream AUC under weak supervision is bounded by the VLM teacher's bias floor $\delta$ independently of student capacity (Section~\ref{sec:capacity}), clinical weak-supervision benchmarks should adopt three reporting standards: (i) report gold-only and weak-augmented scaling curves jointly rather than at a single arbitrary $(\ngold, \nweak)$ operating point; (ii) report the VLM's validation ceiling $\widehat{\mathrm{AUC}}_{\mathrm{FM}}$ alongside per-class toxicity flags ($\widehat{\mathrm{AUC}}^c_{\mathrm{FM}} < 0.50$) so safety regressions on healthy-patient classes such as \texttt{No Finding} cannot hide behind macro-averages; and (iii) condition evaluations of noise-mitigation interventions (confidence filtering, GCE, Co-teaching) on whether the operating point lies below or above $\ngold^\star$.

\paragraph{Limitations and Future Directions.}
First, our empirical grids evaluate in-distribution train/test splits within each benchmark; characterizing how cross-hospital distribution shift between the gold validation set and the unlabeled clinical pool perturbs the decision margin $M(\ngold)$ is an important next step. Second, while we evaluate two VLM labelers (BiomedCLIP across four prompt templates and PubMedCLIP) and six student architectures, extending the multi-VLM comparison to pathology-specialized foundation models such as PLIP \citep{huang2023visual} and CONCH \citep{lu2024conch} or multi-VLM consensus ensembles \citep{ratner2016data,zhong2024vlmcpl} will test how ensemble disagreement reduces error correlation $\rho_{\mathrm{err}}$. Third, our student experiments train decoupled classifiers; extending the crossover analysis to parameter-efficient fine-tuning (e.g., LoRA or linear probing) of the VLM backbone itself remains an open direction.

\paragraph{Conclusion.}
VLM-generated weak labels are a powerful variance-reduction scaffold when expert annotations are scarce, but they impose a rigid, architecture-invariant bias ceiling once gold supervision suffices to match the VLM. By calibrating this noisy-label crossover across three clinical imaging modalities, six student architectures, and 15 thoracic pathologies---and distilling the theory into an audited, zero-fitting decision rule operable from a small validation split---this work provides clinical AI practitioners with a rigorous, actionable criterion for when to leverage foundation-model pseudo-labels and when to invest in expert clinical annotation.

\section*{Acknowledgments}
We thank colleagues at Stanford University and Mayo Clinic for insightful discussions and feedback throughout the development of this work. A condensed version of this work was accepted for presentation at the ICML 2026 Workshop on Combining Theory and Benchmarks (CTB).

\bibliographystyle{unsrtnat}
\bibliography{refs}

\end{document}